\documentclass[lettersize,journal]{IEEEtran}
\usepackage{amsmath,amsfonts}
\usepackage{algorithmic}
\usepackage{algorithm}
\usepackage{array}
\usepackage[caption=false,font=normalsize,labelfont=sf,textfont=sf]{subfig}
\usepackage{textcomp}
\usepackage{stfloats}
\usepackage{url}
\usepackage{verbatim}
\usepackage{graphicx}
\usepackage{cite}

\newcommand{\hzz}[1]{{\color{black}{#1}}}
\usepackage{booktabs}
\usepackage{mathrsfs}
\usepackage{multirow}
\usepackage{color}
\usepackage{colortbl}
\usepackage{arydshln}
\usepackage{array}
\usepackage{bm}
\usepackage{rotating}
\usepackage{adjustbox}
\usepackage{color}

\usepackage{xcolor}
\definecolor{Gray}{gray}{0.9}

\begin{document}

\title{A Generic Shared Attention Mechanism for Various Backbone Neural Networks}

	\author{Zhongzhan~Huang,
		Senwei Liang,
		Mingfu Liang,
		Liang Lin$^{\dagger}$\thanks{$^{\dagger}$Corresponding authour. }, \textit{Fellow, IEEE}
		\IEEEcompsocitemizethanks{\IEEEcompsocthanksitem Z. Huang and L. Lin are with the School of Computer Science and Engineering, Sun Yat-sen University, Guangzhou, China (e-mail: linliang@ieee.org). 
		S. Liang is with Department of Mathematics, Purdue University, West Lafayette, IN, USA. 
		M. Liang is with the Department of Electrical and Computer Engineering, Northwestern University, Evanston, IL, USA.

  A preliminary version of this work has been presented in the AAAI \cite{huang2020dianet}.
		}
		}

\markboth{Preprint}%
{Shell \MakeLowercase{\textit{et al.}}: Preprint}

\maketitle

\begin{abstract}
The self-attention mechanism has emerged as a critical component for improving the performance of various backbone neural networks. However, current mainstream approaches individually incorporate newly designed self-attention modules (SAMs) into each layer of the network for granted without fully exploiting their parameters' potential. This leads to suboptimal performance and increased parameter consumption as the network depth increases. To improve this paradigm, in this paper, we first present a counterintuitive but inherent phenomenon: SAMs tend to produce strongly correlated attention maps across different layers, with an average Pearson correlation coefficient of up to 0.85. Inspired by this inherent observation, we propose Dense-and-Implicit Attention (DIA), which directly shares SAMs across layers and employs a long short-term memory module to calibrate and bridge the highly correlated attention maps of different layers, thus improving the parameter utilization efficiency of SAMs. This design of DIA is also consistent with the neural network's dynamical system perspective. Through extensive experiments, we demonstrate that our simple yet effective DIA can consistently enhance various network backbones, including ResNet, Transformer, and UNet, across tasks such as image classification, object detection, and image generation using diffusion models. Furthermore, our further comprehensive analysis reveals that the effectiveness of DIA primarily arises from its dense and implicit inter-layer information connections, which are absent in conventional self-attention mechanisms. These connections not only stabilize the neural network training but also provide regularization effects similar to well-known techniques like skip connections and batch normalization. The insights in this paper enhance our understanding of attention mechanisms, facilitating their optimization and laying a solid foundation for future advancements and broader applications across diverse backbone neural networks.
\end{abstract}

\begin{IEEEkeywords}
Layer-wise shared Attention mechanism, Parameter sharing, Dense-and-Implicit connection, Stable training.
\end{IEEEkeywords}

\section{Introduction}
\label{sec:introduction}
\begin{figure}[t]
	 		\centering
	 		\includegraphics[width=0.9\linewidth]{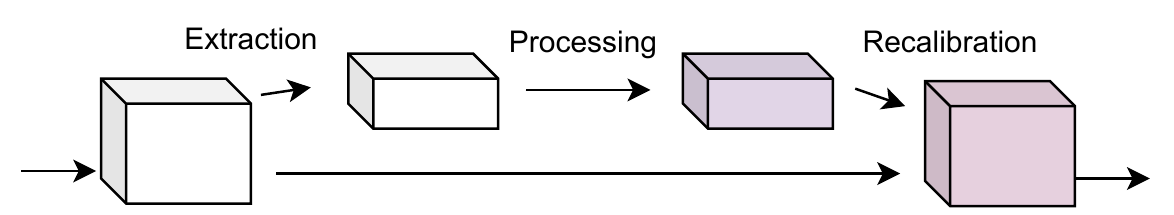}
	 				\vspace{-0.1cm}
	 		\caption{The paradigm of self-attention module~(SAM).} 
	 \vspace{-0.2cm}
	 		\label{fig:net}
	 		\vspace{-0.3cm}
	 	\end{figure}

\IEEEPARstart{A}{ttention}, a cognitive process that selectively focuses on a small part of information while neglects other perceivable information~\cite{anderson2005cognitive,xie2021gpca}, has been used to effectively ease neural networks from learning large information contexts from sentences~\cite{vaswani2017attention,britz2017massive,cheng2016long}, images~\cite{Xu:2015:SAT:3045118.3045336,luong2015effective} and videos~\cite{miech2017learnable}. For example in computer vision, deep neural networks~(DNNs) incorporated with special operators that mimic the attention mechanism can process informative regions of an image efficiently. These operators are modularized and plugged into networks as attention modules~\cite{hu2018squeeze,woo2018cbam,park2018bam,wang2018non,hu2018gather,cao2019GCNet}.
Generally, the self-attention modules~\hzz{(SAMs)} can be divided into three parts: ``extraction", ``processing" and ``recalibration" as shown in Fig.~\ref{fig:net}. First, the plug-in module extracts internal features of a network which can be squeezed as the channel-wise information~\cite{hu2018squeeze,li2019selective}, spatial information~\cite{wang2018non,woo2018cbam,park2018bam}, etc. Next, the module processes the extraction and generates an attention map to measure the importance of the features map via a fully connected layer~\cite{hu2018squeeze}, convolution layer~\cite{wang2018non}, etc. Last, the attention map is applied to recalibrate the features. 
Most existing works adopt a common paradigm where the attention modules are individually plugged into each block throughout DNNs~\cite{hu2018squeeze,woo2018cbam,park2018bam,wang2018non}. This paradigm facilitates researchers to focus on the optimization of three parts in the SAM 
and 
allows a fair and fast comparison of the performance of the designed modules, which has greatly facilitated the community. 
   
	 	 	\begin{figure*}[t]
	 		\centering
	 		\includegraphics[width=0.9\textwidth]{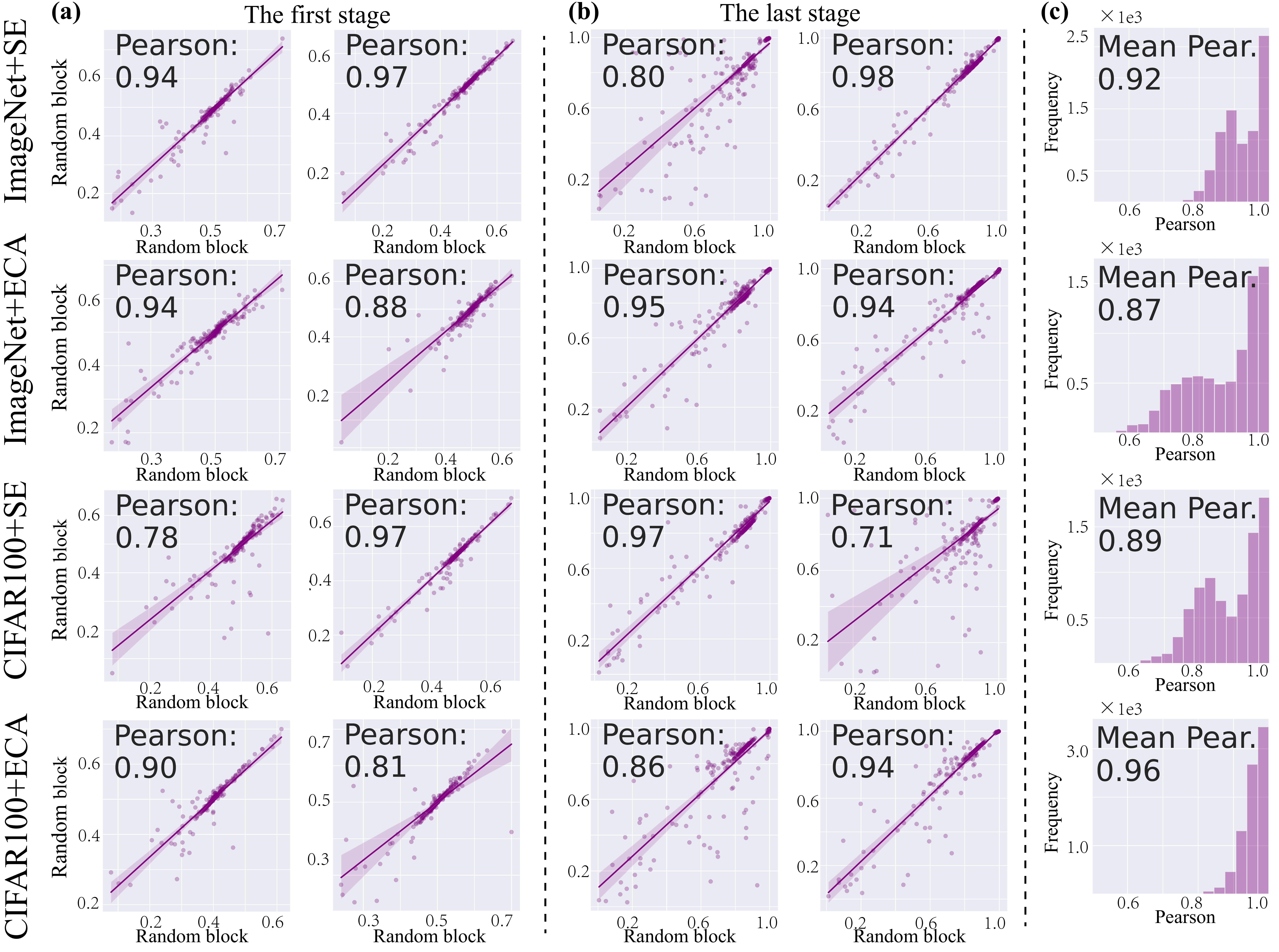}
	 				\vspace{-0.5cm}
	 		\caption{\hzz{Linear correlation of attention maps across different layers. Experiments on ImageNet and CIFAR100 are conducted using ResNet50 and ResNet164 as backbones, respectively. The attention modules ``SE" and ``ECA" refer to those from two well-known channel attention networks, SENet \cite{hu2018squeeze}, and ECANet \cite{wang2020eca}. \textbf{(a)} and \textbf{(b)} respectively represent the linear relationship of attention maps generated in two randomly sampled blocks in the first stage and the last stage. The visualization of individual input samples is provided by \textbf{(a)} and \textbf{(b)}, while \textbf{(c)} displays the distribution of Pearson correlation coefficients across multiple input samples in different pair of blocks.}} 
	 \vspace{-0.2cm}
	 		\label{fig:pearson}
	 	\end{figure*}

However, the drawbacks of such a paradigm are obvious, i.e., the parameter and memory cost create a bottleneck with the increasing numbers of SAMs. Specifically, for DNNs, the performance and depth are closely related. Under a suitable training setting, deeper networks can achieve better performance \cite{zhong2022mix,zhong2022switchable}. This implies that if we aim to enhance a deep network by SAMs and individually connect the SAMs to each block of the entire DNN backbone, it will inevitably result in an increasing number of parameters as the network depth increases. 
Therefore, we urgently need an efficient and simple approach to optimize the utilization of parameters for SAMs within existing paradigms, aiming to improve the model's performance and mitigate parameter wastage. 
In this paper,  we propose a novel-and-straightforward attention mechanism, called Dense-and-Implicit-Attention (DIA), to alleviate the aforementioned drawback and achieve a performance improvement, inspired by a counterintuitive but inherent phenomenon: \textbf{SAMs tend to produce strongly correlated attention maps across different layers.}

	 	\begin{figure*}[t]
	 		\centering
	 		\includegraphics[width=0.9\textwidth]{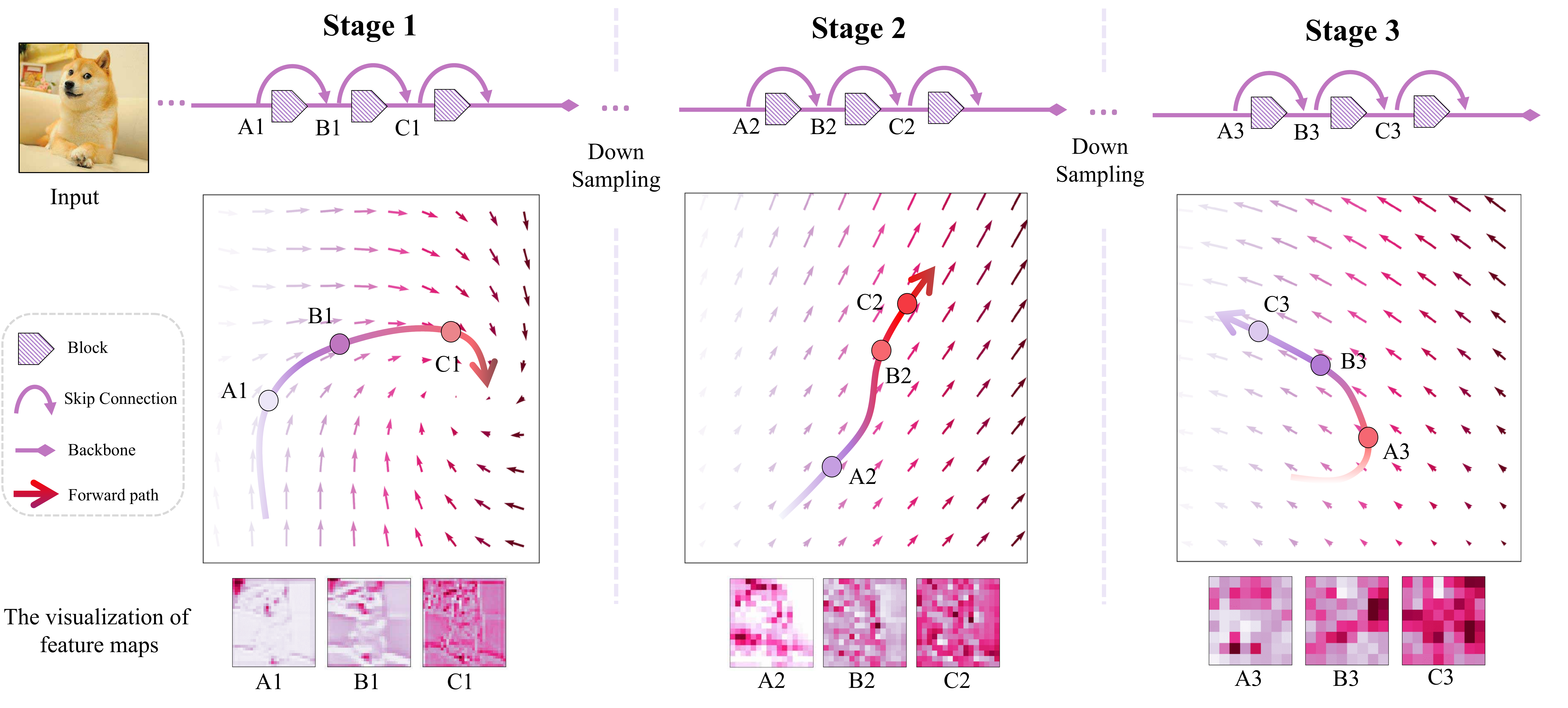}
	 				\vspace{-0.5cm}
	 		\caption{The Residual neural network on dynamical system perspective. The forward process of ResNet can be modeled as a forward numerical method for a dynamical system. The feature map of any block in the same stage can be regarded as a point in the same vector space. For example, in Stage 1, ``A1", ``B1", and ``C1" are the feature maps from three adjacent blocks, and these feature maps can be regarded as homogenous. } 
	 \vspace{-0.2cm}
	 		\label{fig:vector}
	 	\end{figure*}	

	 	 	\begin{figure*}[t]
	 		\centering
	 		\includegraphics[width=0.75\textwidth]{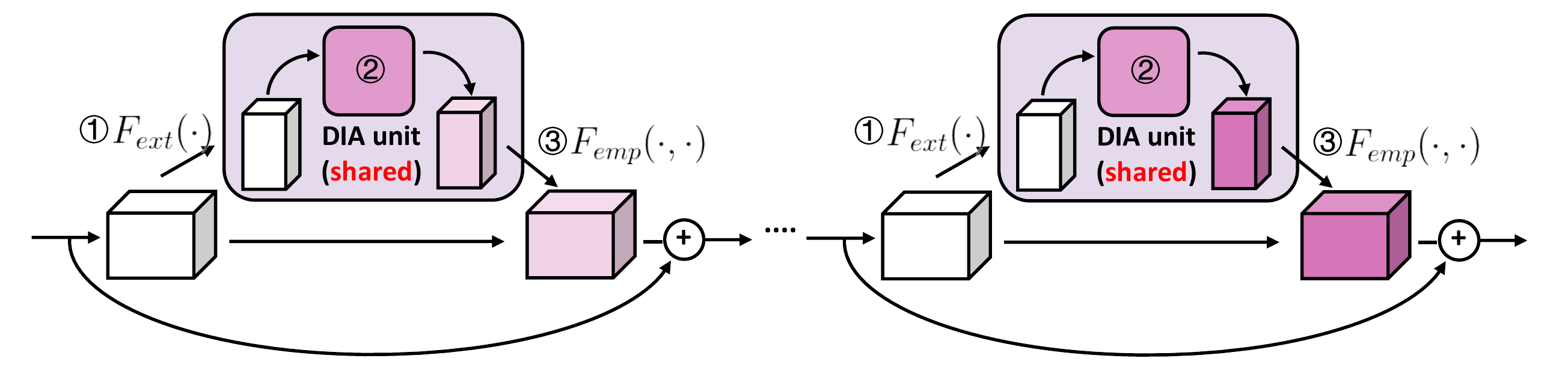}
	 				\vspace{-0.5cm}
	 		\caption{The DIA architecture in the residual network. $F_{ext}$ means the operation for extracting different scales of features. $F_{emp}$ means the operation for emphasizing or recalibrating features.} 
	 \vspace{-0.2cm}
	 		\label{fig:diaframwork}
	 	\end{figure*}

\hzz{Specifically, let's denote the input to the network as $x$, and consider the attention maps generated by SAMs from the $i$-th and $j$-th layers within the same stage, which we'll refer to as $a_i(x)$ and $a_j(x)$, respectively. Since these layers\footnote{The "block" and "layer" are used interchangeably in this paper unless otherwise specified, as different backbone neural networks may have slight variations in naming their components. } belong to the same stage\footnote{The "Stage" is a common component in various backbone neural networks, typically composed of multiple layers or blocks with the same structure. Blocks or layers within the same stage usually yield features of the same dimension.}, they share the same dimension, denoted as $n$. To measure the correlation between the corresponding $n$ elements of $a_i(x)$ and $a_j(x)$, we employ the Pearson correlation coefficient~\cite{rodgers1988thirteen}. If the correlation coefficient is close to 1, it suggests a highly linear correlation between $a_i(x)$ and $a_j(x)$, indicating that there exists a constant $c \in \mathbb{R}$ such that $a_i(x)\approx c\cdot a_j(x)$. Generally, a correlation coefficient greater than 0.8 indicates a strong correlation between the two attention maps.

In our investigation, we utilize ResNet164 and ResNet50 as backbone networks, and Fig. \ref{fig:pearson} visually represents the inter-layer correlations of attention maps across different datasets and SAMs configurations. Specifically, Fig. \ref{fig:pearson} (a-b) showcases the correlations between attention maps from randomly selected layers belonging to distinct stages of the residual network, while Fig. \ref{fig:pearson} (c) provides an overall overview of the Pearson coefficients for pairwise layers. Notably, we consistently observe that the average inter-layer Pearson correlation coefficients exceed 0.85, with a majority clustering around 0.9. Moreover, this phenomenon remains consistent irrespective of the chosen SAMs module, dataset, stage, or network backbone. Consequently, it implies that the variations in attention maps between layers are not substantial, revealing a strong inherent correlation among the attention maps. 
Therefore, placing SAM individually in each block of the backbone network seems unnecessary, and the attention maps can be generated in a unified manner. 

Inspired by this inherent observation, our proposed DIA considers sharing SAMs within the same stage, enabling the calibration and unification of highly correlated attention maps across different layers. Consequently, this approach explicitly and directly enhances the parameter utilization efficiency of SAMs. }
Specifically, the original paradigm of the attention mechanism can be expressed as follows:
\begin{equation}
    x_{t+1} = x_t + \underbrace{f(x_t;\theta_t)}_{\text{Feature map}} \otimes \underbrace{\mathbf{A}(f(x_t;\theta_t);\phi_t)}_{\text{Attention map}},
\label{eq:att}
\end{equation}
where separate SAMs $\mathbf{A}(\cdot;\phi_t)$ are customized for each block with learnable parameters $\phi_t$. $x_t$ is the feature of the $t^{\rm th}$ layer, and $f(\cdot;\theta_t)$ is the neural network in the corresponding layer with learnable parameters $\theta_t$. The proposed DIA first considers a novel paradigm that utilizes a single SAM $\mathbf{A}(\cdot;{\color{purple}\phi_{\text{share}}})$ with shared parameters ${\color{purple}\phi_{\text{share}}}$ instead of individual SAMs $\mathbf{A}(\cdot;\phi_t)$ for each block. i.e.,
\begin{equation}
    x_{t+1} = x_t + \underbrace{f(x_t;\theta_t)}_{\text{Feature map}} \otimes \underbrace{\mathbf{A}(f(x_t;\theta_t);{\color{purple}\phi_{\text{share}}} )}_{\text{Attention map}}.
\label{eq:att_dia}
\end{equation}
Moreover, in order to further enhance the inter-layer information aggregation and calibration of highly correlated attention maps in SAMs, module $\mathbf{A}(\cdot;{\color{purple}\phi_{\text{share}}})$ can be further designed, and it will be discussed in Section \ref{sec:method}.

Indeed, the idea of layer-wise shared attention mechanism DIA, also aligns with the fundamental perspective of residual neural networks as dynamical systems. 
The dynamical system perspective~\cite{weinan2017proposal,queiruga2020continuous,zhu2022convolutional,meunier2022dynamical} is a crucial hypothesis that explains various behaviors of DNNs and offers insights for designing effective DNN structures. A dynamical system can be expressed as follows:
\begin{align}
\frac{\text{d}\mathbf{u}}{\text{d}t} = \mathbf{f}(\mathbf{u}), \quad \mathbf{u}(0) = \mathbf{c}_0,
\label{eqn:ode}
\end{align}
where $\mathbf{c}_0$ represents the initial state. As demonstrated in Table \ref{tab:dyandres}, the forward process of ResNet can be likened to the forward scheme used to solve a dynamic system. The residual block $f(\cdot;\theta_t)$ in the $t^{\rm th}$ block, can be seen as an integration $S(\cdot;\mathbf{f},\Delta t)$ using a numerical scheme $S$ with a step size $\Delta t$. Based on this analogy, certain existing studies~\cite{wang2021recurrent,stelzer2021deep,Jaiswal2021TDAMTA,shen2021sliced,zhang2022minivit} propose that since the numerical methods consistently employ the same integration scheme $S$ to solve Eq.~(\ref{eqn:ode}) for the state $\mathbf{u}(t)$, they can likewise use the residual block with the shared parameters to process the input $x_t$ in ResNet. \hzz{Following this perspective, as depicted in Fig.~\ref{fig:vector}, the output of the block within the same stage (a group of consecutive layers with identical spatial dimensions) evolves within the same vector space and these outputs can be intuitively considered homogeneous.} Note that the output of the residual block is also the input of the SAM, and hence we may use a single shared SAM to process the information from the same vector space.

\begin{table}[t]
  \centering
  \caption{Formulation of forward scheme of dynamical system and forward process of ResNet.}
  \vspace{-0.1cm}
    \begin{tabular}{ll}
    \toprule
    \multicolumn{1}{c}{\textbf{Forward scheme of dynamics}} & \multicolumn{1}{c}{\textbf{Forward process of ResNet}} \\
    \midrule
    $\mathbf{u}_{t+1} = \mathbf{u}_t + S(\mathbf{u}_t; \mathbf{f},\Delta t)$     & $x_{t+1} = x_t + f(x_t;\theta_t)$  \\
    \bottomrule
    \end{tabular}%
  \label{tab:dyandres}%
\end{table}%

	 	\begin{figure*}[t]
	 		\centering
	 		\includegraphics[width=0.8\textwidth]{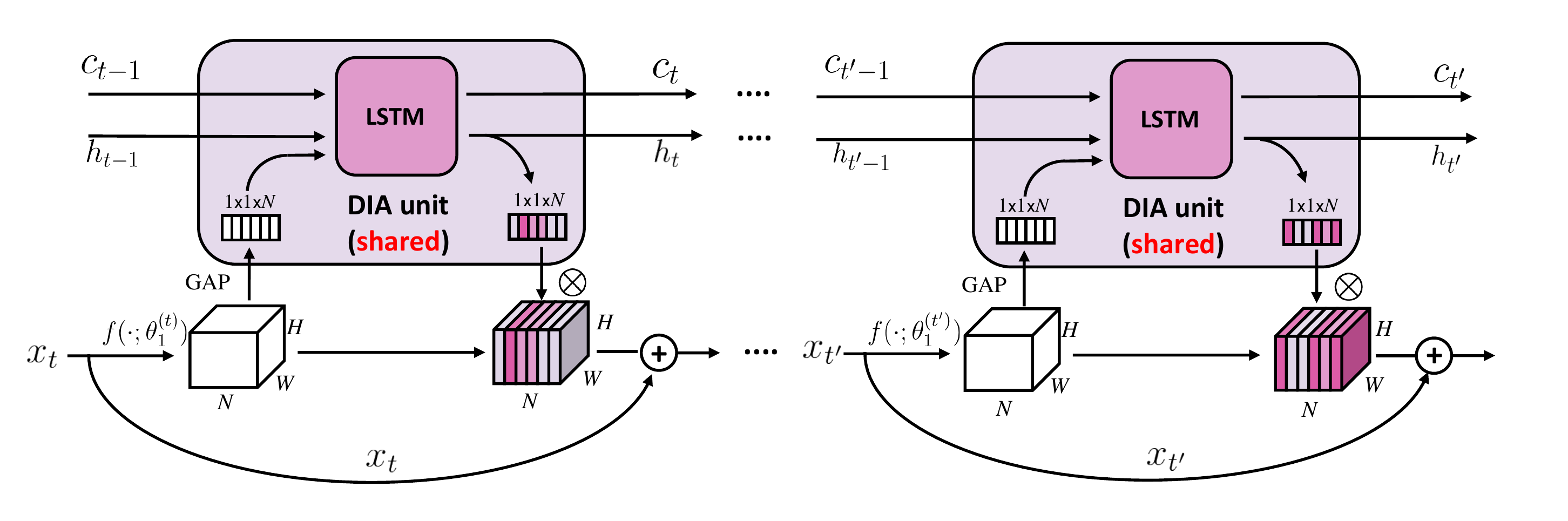}
	 				\vspace{-0.7cm}
	 		\caption{The showcase of DIA-LSTM in detail. In the LSTM cell, $c_t$ is the cell state vector and $h_t$ is the hidden state vector. GAP means global average pool over channels and $\otimes$ means channel-wise multiplication.} 
	 \vspace{-0.2cm}
	 		\label{fig:diaunit}
	 	\end{figure*}

In this paper, we demonstrate that our simple yet effective DIA can \hzz{consistently enhance various network backbones, including ResNet, Transformer, and UNet, across tasks such as image classification, object detection, and image generation using diffusion models}. Moreover, we discover that DIA establishes dense and implicit connections between the self-attention modules and the backbone network. We also find that this integration can serve as a beneficial regularizer that effectively stabilizes the training of the neural network, similar to the well-known techniques like skip connection and batch normalization. To demonstrate the robustness of our approach, we intentionally destabilize the model training by: (1) removing the skip connection of the ResNet, (2) eliminating batch normalization from the model, and (3) omitting all data augmentation during training. Despite these destabilizing factors, DIA still demonstrates stable performance, highlighting its effectiveness and resilience.

\subsection{Our contribution}
	We summarize our contribution as follows,
	\begin{enumerate}
        \item \hzz{Inspired by a counterintuitive but inherent phenomenon we observed, i.e., the strong correlation in attention maps, we propose a generic shared attention mechanism called the DIA to improve the parameters' utilization of SAMs.} 
	    \item We demonstrate the effectiveness of DIA in various vision tasks through experiments conducted on various datasets and popular network architectures. Besides, we present a highly lightweight of DIA, which significantly reduces the parameter cost compared to the original DIA.
	    \item Through extensive experimentation, we discover that our approach can establish dense and implicit connections between the self-attention modules and the backbone network. This integration acts as a good regularizer to effectively stabilize the neural network training.
	\end{enumerate}

\section{Related Works} 
	\label{relatedwork}
	\subsection{Attention Mechanism in Computer Vision.} Previous works use the attention mechanism in image classification via utilizing a recurrent neural network to select and process local regions at high-resolution sequentially~\cite{mnih2014recurrent,zhao2017diversified,huang2022lottery,huang2020efficient}. Concurrent attention-based methods tend to construct operation modules to capture non-local information in an image~\cite{wang2018non,cao2019GCNet}, and model the interrelationship between channel-wise features~\cite{hu2018squeeze,hu2018gather}. The combination of multi-level attention is also widely studied~\cite{park2018bam,woo2018cbam,DBLP:journals/corr/abs-1904-04402,Wang_2017_CVPR}. Prior works~\cite{li2019spatial,wang2018non,liang2020instance,hu2018squeeze,hu2018gather,park2018bam,woo2018cbam,DBLP:journals/corr/abs-1904-04402} usually insert an attention module in each layer individually. In this work, the DIA unit is innovatively shared for all the layers in the same stage of the network, and the existing attention modules can be composited into the DIA unit readily. Besides, we adopt a global average pooling in part \raisebox{.9pt}{\textcircled{\raisebox{-.9pt}{1}}} to extract global information and a channel-wise multiplication in part \raisebox{.9pt}{\textcircled{\raisebox{-.9pt}{3}}} to recalibrate features, which is similar to SENet~\cite{hu2018squeeze}. 
	\begin{table*}[t]
		\centering
		\caption{Formulation for the structure of ResNet, SENet, and DIA-LSTM. $f$ is the neural network in $t^{\rm th}$ block with learnable parameters $\theta_t$. $\text{GAP}(\cdot)$ means global average pooling. $\text{FC}$ means fully connected layer, and $\phi_{t}$ is the trainable parameter within the SE module at the $t^{\rm th}$ layer.}
		\begin{tabular}{|c|c|c|c|}
			\hline
			&    ResNet   &  SENet~\cite{hu2018squeeze}     & DIA-LSTM (ours) \\
			\hline
			(a)  &  $a_t = f(x_t;\theta_{t})$ & $a_t = f(x_t;\theta_{t})$ & $a_t = f(x_t;\theta_{t})$  \\
			(b)  &    -   &   $h_t = \text{FC}(\text{GAP}(a_t);\phi_t)$    &  $(h_t,c_t) = \text{LSTM}(\text{GAP}(a_t), h_{t-1},c_{t-1};\phi_{\text{share}})$\\
			(c) & $x_{t+1} = x_t + a_t$      &    $x_{t+1} = x_t + a_t \otimes h_t$   & $x_{t+1} = x_t + a_t \otimes h_t$ \\
			\hline
		\end{tabular}%

		\label{DIANet compare SENet}%
	\end{table*}%
\subsection{Dense Network Topology.}
	DenseNet proposed in  \cite{huang2017densely} connects all pairs of layers directly with an identity map. Through reusing features, DenseNet has the advantage of higher parameter efficiency, a better capacity of generalization, and more accessible training than alternative architectures~\cite{zhao2021recurrence,lin2013network,he2016deep,srivastava2015highway}. Instead of explicitly dense connections, the DIA unit implicitly links layers at different depths via a shared module and leads to dense connections.
	
\subsection{Multi-Dimension Feature Integration.}
	L. Wolf et al\cite{wolf2006critical} experimentally analyze that even the simple aggregation of low-level visual features sampled from a wide inception field can be efficient and robust for context representation, which inspires \cite{hu2018squeeze,hu2018gather} to incorporate multi-level features to improve the network representation. H. Li et al\cite{li2016multi} also demonstrate that by biasing the feature response in each convolutional layer using different activation functions, the deeper layer could achieve a better capacity for capturing the abstract pattern in DNN. In the DIA unit, the high non-linearity relationship between multidimensional features is learned and integrated via the LSTM module.

\begin{table}[htbp]
  \centering
  \caption{Comparison of performance between existing SAMs incorporated with and without our DIA framework on CIFAR10/100 dataset. ``Org'' represents the backbone without any self-attention module. The p-value ($p$) indicates the results of the Student's t-test, comparing the accuracies of DIA-LSTM with other models. The significance level $(\alpha)$ is set at 0.05. Notation: `` * '': $p < 0.05$. The best results for each backbone are highlighted in bold. The accuracy improvement is highlighted in {\color{red}{red}}, while performance degradation is indicated in {\textcolor{teal}{green}}. } 
\resizebox*{0.99\linewidth}{!}{
     \begin{tabular}{lllccc}
    \toprule
    \multicolumn{1}{c}{{Model}} & \multicolumn{1}{c}{{Dataset}} & \multicolumn{1}{c}{{Backbone}} & \multicolumn{1}{c}{{ \hzz{Acc. \textbf{w/o DIA}} }} & \multicolumn{1}{c}{{ \hzz{Acc. \textbf{with DIA}} }}  & \multicolumn{1}{c}{{p-value}} \\
    \midrule
    Org & CIFAR10 & ResNet83 & 93.62$\pm$ 0.20      &  -        & N/A \\
    SE~\cite{hu2018squeeze} & CIFAR10 & ResNet83 & 94.20$\pm$ 0.16      & \pmb{94.32$\pm$ 0.11} ({\color{red}{$\uparrow$ 0.12}})            &  *\\
    ECA~\cite{wang2020eca}  & CIFAR10 & ResNet83 & 93.32$\pm$ 0.23      & \pmb{93.58$\pm$ 0.15} ({\color{red}{$\uparrow$ 0.26}})             & * \\
    CBAM~\cite{woo2018cbam}  & CIFAR10 & ResNet83 &  \pmb{93.31$\pm$ 0.32}     &  93.01$\pm$ 0.44 ({\textcolor{teal}{$\downarrow$ 0.30}})          & * \\
    SRM~\cite{lee2019srm}   & CIFAR10 & ResNet83 & 94.22$\pm$ 0.14      & \pmb{94.41$\pm$ 0.13} ({\textcolor{red}{$\uparrow$ 0.19}})           & * \\
    SPA~\cite{guo2020spanet}   & CIFAR10 & ResNet83 &  94.52$\pm$ 0.14     & \pmb{94.81$\pm$ 0.22} ({\textcolor{red}{$\uparrow$ 0.29}})            & * \\
    \midrule

    Org & CIFAR10 & ResNet164 & 93.16$\pm$ 0.15      &  -        & N/A \\
    SE~\cite{hu2018squeeze} & CIFAR10 & ResNet164 & 94.32$\pm$ 0.20      &  \pmb{94.62$\pm$ 0.11} ({\color{red}{$\uparrow$ 0.30}})         & * \\
    ECA~\cite{wang2020eca}  & CIFAR10 & ResNet164 & 94.47$\pm$ 0.12      & \pmb{94.87$\pm$ 0.17} ({\color{red}{$\uparrow$ 0.40}})           & * \\
    CBAM~\cite{woo2018cbam}  & CIFAR10 & ResNet164 & 93.82$\pm$ 0.11      &  \pmb{94.63$\pm$ 0.05} ({\color{red}{$\uparrow$ 0.81}})        & * \\
    SRM~\cite{lee2019srm}   & CIFAR10 & ResNet164 & 94.52$\pm$ 0.21      &  \pmb{94.75$\pm$ 0.10} ({\color{red}{$\uparrow$ 0.24}})         & * \\
    SPA~\cite{guo2020spanet}   & CIFAR10 & ResNet164 & 94.22$\pm$ 0.14      & \pmb{94.76$\pm$ 0.10} ({\color{red}{$\uparrow$ 0.54}})             & * \\
    \midrule
    Org & CIFAR100 & ResNet83 & 73.55$\pm$ 0.34      &  -         & N/A \\
    SE~\cite{hu2018squeeze} & CIFAR100 & ResNet83 & 74.68$\pm$ 0.36      &  \pmb{74.89$\pm$ 0.38} ({\color{red}{$\uparrow$ 0.21}})           & * \\
    ECA~\cite{wang2020eca}  & CIFAR100 & ResNet83 & 74.54$\pm$ 0.37      &  \pmb{74.75$\pm$ 0.09} ({\color{red}{$\uparrow$ 0.21}})           &  *\\
    CBAM~\cite{woo2018cbam}  & CIFAR100 & ResNet83 &   73.24$\pm$ 0.20      &  \pmb{73.83$\pm$ 0.25} ({\color{red}{$\uparrow$ 0.59}})          & * \\
    SRM~\cite{lee2019srm}   & CIFAR100 & ResNet83 &    74.48$\pm$ 0.33      &  \pmb{74.70$\pm$ 0.25} ({\color{red}{$\uparrow$ 0.22}})          & * \\
    SPA~\cite{guo2020spanet}   & CIFAR100 & ResNet83 &   75.18$\pm$ 0.25      &  \pmb{75.68$\pm$ 0.28} ({\color{red}{$\uparrow$ 0.50}})            & * \\
    \midrule
    Org & CIFAR100 & ResNet164 & 74.32$\pm$ 0.22      &  -        & N/A \\
    SE~\cite{hu2018squeeze} & CIFAR100 & ResNet164 & 75.23$\pm$ 0.18      & \pmb{75.95$\pm$ 0.15} ({\color{red}{$\uparrow$ 0.62}})           & * \\
    ECA~\cite{wang2020eca}  & CIFAR100 & ResNet164 &  74.15$\pm$ 0.36      & \pmb{74.57$\pm$ 0.25} ({\color{red}{$\uparrow$ 0.42}})          & * \\
    CBAM~\cite{woo2018cbam}  & CIFAR100 & ResNet164 & 73.68$\pm$ 0.21      & \pmb{73.82$\pm$ 0.14} ({\color{red}{$\uparrow$ 0.14}})           &  *\\
    SRM~\cite{lee2019srm}   & CIFAR100 & ResNet164 &  \pmb{74.95$\pm$ 0.67}     & 74.69$\pm$ 0.64 ({\textcolor{teal}{$\downarrow$ 0.26}})           & * \\
    SPA~\cite{guo2020spanet}   & CIFAR100 & ResNet164 & 75.36$\pm$ 0.28      & \pmb{75.93$\pm$ 0.34} ({\color{red}{$\uparrow$ 0.57}})            & *\\
    \bottomrule
    \end{tabular}%
    }
  \label{tab:diaformodule}%
\end{table}%

	 	\begin{figure*}[t]
	 		\centering
	 		\includegraphics[width=1\textwidth]{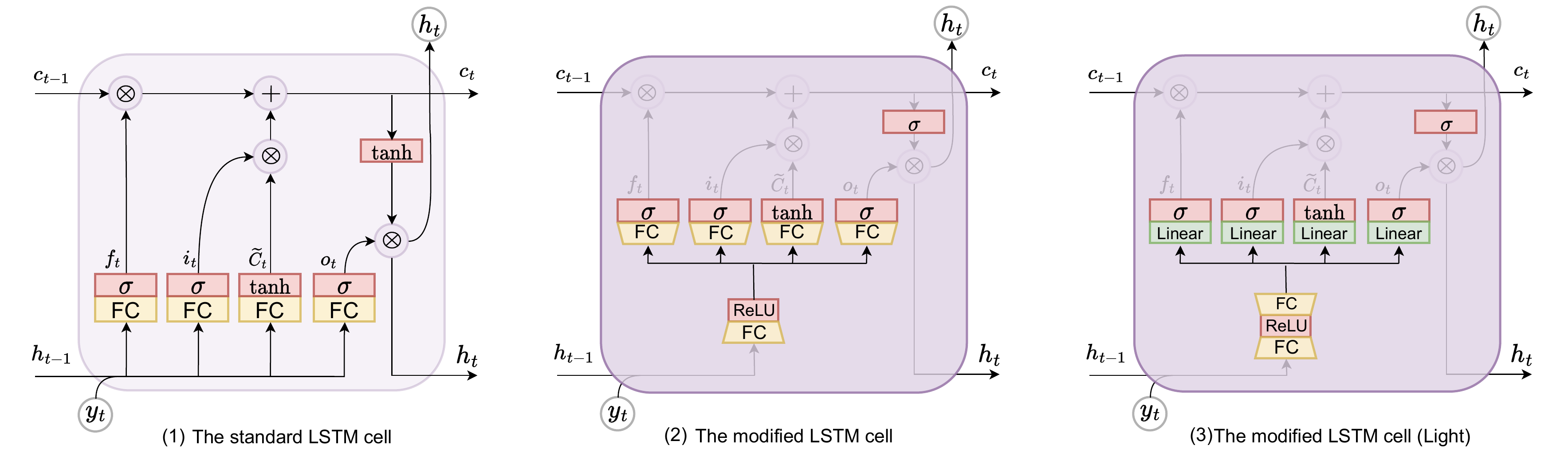}
	 				\vspace{-0.7cm}
	 		\caption{\textbf{Left}. The standard LSTM cell. \textbf{Middle}. The modified LSTM cell is used in DIA-LSTM. We highlight the modified component in the modified LSTM. The symbol "$\sigma$" denotes the sigmoid activation, while "FC" refers to the fully connected layer. \textbf{Right}. A lightweight LSTM cell is used in DIA-LSTM (Light). "Linear" represents a straightforward element-wise linear transformation, as expressed in Sec.~\ref{sec:ie} } 
	 \vspace{-0.2cm}
	 		\label{fig:lstm}
	 	\end{figure*}

\section{Dense-and-Implicit-Attention}
\label{sec:method}

According to Eq.(\ref{eq:att_dia}), the structure and computation flow of incorporating the shared attention mechanism with a SAM is visualized in Fig.~\ref{fig:diaframwork}. There are also three parts: extraction~(\raisebox{.9pt}{\textcircled{\raisebox{-.9pt}{1}}}), processing~(\raisebox{.9pt}{\textcircled{\raisebox{-.9pt}{2}}}) and recalibration~(\raisebox{.9pt}{\textcircled{\raisebox{-.9pt}{3}}}) in DIA. The \raisebox{.9pt}{\textcircled{\raisebox{-.9pt}{2}}} is the main module in DIA to model network attention and is the key innovation of the proposed method where the parameters of the attention module are shared across different layers.

In this section, we begin by confirming the effectiveness of our proposed layer-wise shared-attention mechanism, known as the DIA unit, for existing self-attention modules. Subsequently, we introduce our proposal of integrating Long Short-Term Memory (LSTM)~\cite{hochreiter1997long} into the DIA unit and offer insightful design considerations to reduce the parameter requirements of the standard LSTM, which we refer to as DIA-LSTM.

\subsection{\hzz{Incorporating DIA unit with different SAMs}}
\label{sec:diaunit_for_different_module}

First, we examine the compatibility of different SAMs with the DIA unit. We conduct experiments on the CIFAR10/CIFAR100 datasets using popular SAMs, inclduing SE~\cite{hu2018squeeze}, ECA~\cite{wang2020eca}, CBAM~\cite{woo2018cbam}, SRM~\cite{lee2019srm}, and SPA~\cite{guo2020spanet}. Our objective is to evaluate the performance of integrating these SAMs with the DIA unit and assess their effectiveness in reducing the additional parameter cost. The backbone models used are ResNet83 and ResNet164, following the training configuration specified in the CIFAR experiments by~\cite{he2016deep}. The results are presented in Table \ref{tab:diaformodule}, where all models undergo 15 repeated verifications with random seeds. The reported values include the mean classification accuracies and their corresponding standard deviations.

\textbf{Reducing extra parameter cost.} \hzz{Due to the shared mechanism in DIA, regardless of the type of self-attention modules used, if there are $n$ modules shared in one stage, the extra number of parameters from self-attention modules generally decreases to $1/n$ of its original value. Specifically, the number of parameters in each kind of module can be reduced by 94.4\% and 88.9\% with ResNet164 and ResNet83, respectively. However, for SRM (see Section \ref{sec:limit}), since batch normalization is not shared in the stage, the number of SRM parameters can merely be reduced by approximately 40\%. Therefore, the proposed DIA can explicitly and efficiently enhance the utilization of parameters for various SAMs.}

\textbf{Maintaining accuracy.} \hzz{The classification accuracy results in Table \ref{tab:diaformodule} demonstrate that most of the SAMs integrated with DIA exhibit a statistically significant improvement in accuracy (p-value $<0.05$). However, a few SAMs may experience a slight performance loss under certain settings. It is important to note that some of these SAMs were not originally designed specifically for the CIFAR datasets. Nevertheless, in Table \ref{tab:diaformodule}, we reproduce these modules on CIFAR10/CIFAR100, and while the performance of some modules may not be very good, it still effectively illustrates the effectiveness of DIA.}

\subsection{DIA with Long Short-Term Memory (DIA-LSTM)}
\label{sec:DIA-LSTM}

The idea of parameter sharing is also used in recurrent neural networks (RNNs) to capture contextual information. To model layer-wise interrelation, we extend our framework to incorporate RNN. Given that Long Short-Term Memory (LSTM)~\cite{hochreiter1997long} is effective in capturing long-distance dependencies, we primarily focus on utilizing LSTM within the DIA unit (referred to as DIA-LSTM) for the remainder of this paper.
	
    Fig.~\ref{fig:diaunit} is the showcase of DIA-LSTM. A global average pooling~(GAP) layer (as the {\small{\textcircled{\tiny{1}}}} in Fig.~\ref{fig:diaframwork}) is used to extract global information from the current layer. An LSTM module (as the {\small{\textcircled{\tiny{2}}}} in Fig.~\ref{fig:diaframwork}) is used to integrate multi-scale information and there are three inputs passed to the LSTM: the extracted global information from current raw feature map, the hidden state vector $h_{t-1}$, and cell state vector $c_{t-1}$ from previous layers. Then the LSTM outputs the new hidden state vector $h_{t}$ and the new cell state vector $c_{t}$. The cell state vector $c_{t}$ stores the information from the $t^{th}$ layer and its preceding layers. The new hidden state vector $h_{t}$~(dubbed as attention map in our work) is then applied back to the raw feature map by channel-wise multiplication (as the {\small{\textcircled{\tiny{3}}}} in Fig.~\ref{fig:diaframwork}) to recalibrate the feature. 
	The LSTM in the DIA unit plays a role in bridging the current layer and preceding layers such that the DIA unit can adaptively learn the non-linearity relationship between features in two distinct aspects. The first dimension of features is the internal information of the current layer. The second dimension represents the outer information, regarded as layer-wise information, from the preceding layers. The non-linearity relationship between these two dimensions will benefit attention modeling for the current layer.

	\subsubsection{Formulation of DIA-LSTM}\label{subsec:Formulation of DIANet}

As shown in Fig.~\ref{fig:diaunit} when DIA-LSTM is built with a residual network~\cite{he2016deep}, the input of the $t^{\rm th}$ layer is $x_t\in \mathbb{R}^{W\times H \times N}$, where $W, H$ and $N$ mean width, height and the number of channels, respectively. $f(\cdot; \theta_t)$ is the residual mapping  at the $t^{\rm th}$ layer with parameters $\theta_t$ as introduced in~\cite{he2016deep}. Let $a_t = f(x_t; \theta_t)\in\mathbb{R}^{W\times H \times N}.$ Next, a global average pooling denoted as $\text{GAP}(\cdot)$ is applied to $a_t$ to extract global information from features in the current layer. Then $\text{GAP}(a_t)\in \mathbb{R}^{ N}$ is passed to LSTM along with a hidden state vector $h_{t-1}$ and a cell state vector $c_{t-1}$ ( $h_0$ and $c_0$ are initialized as zero vectors). The LSTM finally generates a current hidden state vector $h_{t}\in \mathbb{R}^{N}$ and a cell state vector $c_{t}\in \mathbb{R}^{N}$ as
	\begin{align}
	(h_t,c_t) = \text{LSTM}(\text{GAP}(a_t), h_{t-1},c_{t-1};\phi_{\text{share}}).
	\label{eqn:lstm-dia}
	\end{align}
	 In our model, the hidden state vector $h_t$ is regarded as an attention map to recalibrate feature maps adaptively. We apply channel-wise multiplication $\otimes$ to enhance the importance of features, i.e., $a_t \otimes h_t$ and obtain $x_{t+1}$ after skip connection, i.e., $x_{t+1} = x_t + a_t \otimes h_t$. 
  
  Table~\ref{DIANet compare SENet} shows the formulation of ResNet, SENet~\cite{hu2018squeeze}, and DIA-LSTM, and Part (b) is the main difference between them. The LSTM module is used repeatedly and shared with different layers in parallel to the network backbone. Therefore the number of parameters $\phi_{\text{share}}$ in an LSTM does not depend on the number of layers in the backbone, e.g., $t$. SENet utilizes an attention module consisting of fully connected layers to model the channel-wise dependency for each layer individually~\cite{hu2018squeeze}. The total number of parameters brought by the add-in modules depends on the number of layers in the backbone and increases with the number of layers.

\subsection{Reducing LSTM Module Params in DIA-LSTM}

 In the design of attention modules, it is common to require the attention map values to be within the range of [0,1], representing their importance~\cite{liang2020instance}. Besides, attention modules often aim for a small parameter increment~\cite{li2019spatial}. In the case of the standard LSTM cell, as shown in Figure \ref{fig:lstm} (Left), there are a couple of considerations. Firstly, the activation function used in LSTM, namely tanh, has a range of [-1,1], which differs from the desired [0,1] range for the attention map values. The range of the attention map values has been found to be crucial for the performance of the model~\cite{huang2020dianet}. Secondly, the standard LSTM cell contains several fully connected layers, each with potentially up to $N^2$ learnable parameters, resulting in a significant computational cost.

Therefore, we have made several modifications to the LSTM module used in DIA-LSTM. In Fig.\ref{fig:lstm}, we illustrate the differences between our proposed LSTMs and the standard LSTM~\cite{hochreiter1997long} cell. Specifically, we have incorporated three key modifications in our approach:

\begin{itemize}
    \item A carefully selected activation function;
    \item A shared linear transformation to reduce the input dimension of the LSTM;
    \item A more lightweight element-wise linear transformation.
\end{itemize}

  \subsubsection{Activation Function} The Sigmoid function ($\sigma(z) = 1/(1+e^{-z})$) is widely utilized in various attention-based methods, such as SENet~\cite{hu2018squeeze} and CBAM~\cite{woo2018cbam}, for generating attention maps as a gate mechanism. In our proposed LSTM model, as depicted in Fig.~\ref{fig:lstm}, we have replaced the Tanh activation function in the output layer with the Sigmoid function. We will provide a more detailed analysis and discussion of this modification in the ablation study.
  
\subsubsection{Input Dimension Reduction} A standard LSTM consists of four linear transformation layers as shown in Fig.~\ref{fig:lstm} (Left). Since $y_{t}$, $h_{t-1}$ and $h_{t}$ are of the same dimension $N$, the standard LSTM may cause $8N^2$ parameter increment (The details are shown in Section \ref{sec:paranum}). When the number of channels is large, e.g., $N=2^{10}$, the parameter increment of the added-in LSTM module in the DIA unit will be over 8 million, which can hardly be tolerated.

	As shown in Fig.~\ref{fig:lstm}~(Left), $h_{t-1}$ and $y_{t}$ are passed to four linear transformation layers with the same input and output dimension $N$. In the modified LSTM cell, a linear transformation layer with a smaller output dimension is applied to $h_{t-1}$ and $y_{t}$, where we use reduction ratio $r$ in these transformations. 
	Specifically, we reduce the dimension of the input  from $1 \times 1 \times N$ to $1 \times 1 \times N/r$ and then apply the ReLU activation function to increase non-linearity in this module. The dimension of the output from the ReLU function is changed back to $1 \times 1 \times N$ when the output is passed to those four linear transformation functions. 
	This modification can enhance the relationship between the inputs for different parts in DIA-LSTM and also effectively reduce the number of parameters by sharing a linear transformation for dimension reduction. The number of parameter increments reduces from $8N^2 $ to $10 N^2/r$, and we find that when we choose an appropriate reduction ratio $r$, we can make a better trade-off between parameter reduction and the performance of DIA-LSTM. 

 \subsubsection{DIA-LSTM (Light)}
 \label{sec:ie}
 In Fig.~\ref{fig:lstm} (Right), we propose a modification where some fully connected layers are replaced with an element-wise linear transformation, as described by Eq.~(\ref{eq:ie}). This substitution significantly reduces the number of parameters in DIA-LSTM, but it can potentially result in a performance loss.

For an input vector $x \in \mathbb{R}^n$, the element-wise linear transformation is given by:

\begin{equation}
\mathbf{W}_{\text{Linear}} \otimes x + \mathbf{b}_{\text{Linear}},
\label{eq:ie}
\end{equation}
where $\otimes$ denotes the element-wise multiplication, $\mathbf{W}_{\text{Linear}} \in \mathbb{R}^n$ and $\mathbf{b}_{\text{Linear}} \in \mathbb{R}^n$ are learnable scaling and bias parameters. This lightweight implementation of LSTM is referred to as DIA-LSTM (Light). 

\subsection{Parameter Cost in LSTM and its Variants}
\label{sec:paranum}
This section shows the number of parameter costs in the standard LSTM and the modified LSTM in Fig.~\ref{fig:lstm} with the reduction ratio $r$. The input $y_t$,l
the hidden state vector $h_{t-1}$ and the output in Fig.~\ref{fig:lstm} are of the same size $N$, which is equal to the number of channels.
	
	\noindent\textbf{Standard LSTM.} Since $y_{t}$, $h_{t-1}$, and $h_{t}$ have the same dimensionality $N$, the fully connected layers perform mappings from $\mathbb{R}^N$ to $\mathbb{R}^N$. In the standard LSTM, there are four fully connected layers applied to $y_{t}$ and an additional four fully connected layers applied to $h_{t-1}$. Therefore, a total of eight fully connected layers are involved, resulting in the utilization of $8N^2$ parameters.

	\noindent\textbf{DIA-LSTM.} In Fig.~\ref{fig:lstm} (Middle), an initial fully connected layer is applied to $y_t$ to reduce its dimension from $N$ to $N/r$. This fully connected layer introduces $N^2/r$ parameters. Subsequently, the output is passed through four linear transformations, following the same structure as the standard LSTM. Each of these four fully connected layers accounts for $N^2/r$ parameters, resulting in a total of $4N^2/r$ parameters. Therefore, for the input $y_t$ and reduction ratio $r$, the total number of parameters is $5N^2/r$. Similarly, the number of parameters for the input $h_{t-1}$ is identical to that of the input $y_t$. Thus, the total number of parameters for both inputs is $10N^2/r$.

 \noindent\textbf{DIA-LSTM (Light).} We employ the same fully connected layer as the DIA-LSTM at the outset to process the information, resulting in $N^2/r$ parameters. Subsequently, four element-wise linear transformations are applied, each contributing $2N$ parameters, resulting in a total of $8N$ parameters. Therefore, for the input $y_t$ with a reduction ratio of $r$, the total number of parameters amounts to $N^2/r + 8N$. Similarly, the number of parameters for the input $h_{t-1}$ is identical to that of the input $y_t$. Consequently, the overall number of parameters for both inputs is $2N^2/r + 16N$.

\begin{table}[htbp]
  \centering
  \caption{{Comparison of performance between existing self-attention modules and our DIA-LSTM on CIFAR100 dataset. ``Param'' indicates the number of parameters, with "M" representing the unit of a million. ``Org'' represents the backbone without any self-attention module. The p-value ($p$) indicates the results of the Student's t-test, comparing the accuracies of DIA-LSTM with other models. The significance level $(\alpha)$ is set at 0.05. Notation: `` * '': $p < 0.05$. The best and second-best results for each backbone are highlighted in bold and \underline{\textit{italic}} fonts, respectively. The accuracy improvement is highlighted in {\color{red}{red}}, while performance degradation is indicated in {\textcolor{teal}{green}}.}}
\resizebox*{0.99\linewidth}{!}{
    \begin{tabular}{lllrrcc}
    \toprule
    \textbf{Backbone}  & \textbf{Model} & \multicolumn{1}{c}{\textbf{Param}} & \multicolumn{1}{c}{\textbf{Top1-acc.}} & \multicolumn{1}{c}{\textbf{Top1-acc.↑}} & \multicolumn{1}{c}{\textbf{p-value}} \\
    \midrule
    ResNet164   & Org &  1.73M     &  74.32$\pm$ 0.22     & -      &  *\\
    ResNet164   &  SE~\cite{hu2018squeeze}   &  1.93M     & 75.23$\pm$ 0.18  &      {\color{red}{ 0.91}} &  *\\
    ResNet164   & ECA~\cite{wang2020eca}    &  1.73M     & 74.15$\pm$ 0.36  &  {\textcolor{teal}{ -0.17}}     &  *\\
    ResNet164   & CBAM~\cite{woo2018cbam}  &  1.93M     & 73.68$\pm$ 0.21 &   {\textcolor{teal}{ -0.64}}    &  *\\
    ResNet164   & SRM~\cite{lee2019srm}   &  1.76M     & 74.95$\pm$ 0.67      &  {\color{red}{ 0.63}}     &  *\\
    ResNet164   & SPA~\cite{guo2020spanet}   &  1.96M     & 75.36$\pm$ 0.28      &  {\color{red}{ 1.04}}     & * \\
    ResNet164   & SEM~\cite{zhong2022switchable}   &  1.97M     & 76.36$\pm$ 0.16      &  {\color{red}{ 2.04}}     &  *\\
    ResNet164   & SPEM~\cite{zhong2022mix}   & 1.76M      & 76.26$\pm$ 0.26      & {\color{red}{ 1.94}}      & * \\
    \rowcolor{Gray}ResNet164   & DIA-LSTM &  1.95M     & \underline{\textit{77.09$\pm$ 0.26}}      & {\color{red}{ \underline{\textit{2.77}}}}      &  N/A\\
    \rowcolor{Gray}ResNet164   & DIA-LSTM (Light) &  1.75M     & \pmb{77.48$\pm$ 0.09}      & {\color{red}{ \pmb{3.16}}}     &  *\\
    \midrule
    WRN52-4   & Org &   12.07M    & 79.85$\pm$ 0.21     &  -     & * \\
    WRN52-4   &  SE~\cite{hu2018squeeze}   & 12.42M   & 80.31$\pm$ 0.36 &   {\color{red}{ 0.46}}    & * \\
    WRN52-4   & ECA~\cite{wang2020eca}    &  12.07M     & 79.88$\pm$ 0.38 &  {\color{red}{ 0.03}}     & * \\
    WRN52-4   & CBAM~\cite{woo2018cbam}  &  12.16M     & 79.79$\pm$ 0.63 &   {\textcolor{teal}{ -0.06}}     & * \\
    WRN52-4   & SRM~\cite{lee2019srm}   &  12.09M     & 80.00$\pm$ 0.20      &  {\color{red}{ 0.15}}     & * \\
    WRN52-4   & SPA~\cite{guo2020spanet}   &  13.02M     & 80.11$\pm$ 0.31      &  {\color{red}{ 0.26}}     & * \\
    WRN52-4   & SEM~\cite{zhong2022switchable}   &  12.44M     & 80.36$\pm$ 0.61      &  {\color{red}{ 0.51}}     & * \\
    WRN52-4   & SPEM~\cite{zhong2022mix}   &  12.09M     & 80.72$\pm$ 0.33      &  {\color{red}{0.87}}     & * \\
    \rowcolor{Gray}WRN52-4   & DIA-LSTM &  12.30M     &  \pmb{81.02$\pm$ 0.11}     &  {\color{red}{ \pmb{1.17}}}     & N/A \\
    \rowcolor{Gray}WRN52-4   & DIA-LSTM (Light) & 12.11M       &  \underline{\textit{80.95$\pm$ 0.33}}     &  {\color{red}{\underline{\textit{1.10}}}}    & * \\
    \midrule
    ResNeXt101,8x32   & Org &  32.14M     &    81.09$\pm$ 0.33     &  -     & * \\
    ResNeXt101,8x32   &  SE~\cite{hu2018squeeze}   &  34.03M     & \underline{\textit{82.46$\pm$ 0.31}} &   {\color{red}{\underline{\textit{1.37}}}}    & * \\
    ResNeXt101,8x32   & ECA~\cite{wang2020eca}    &  32.14M    & 81.16$\pm$ 0.28 &   {\color{red}{ 0.07}}    & * \\
    ResNeXt101,8x32   & CBAM~\cite{woo2018cbam}  & 32.61M      & 80.80$\pm$ 0.51 &   {\textcolor{teal}{ -0.29}}    & * \\
    ResNeXt101,8x32   & SRM~\cite{lee2019srm}    &  32.18M     &  81.27$\pm$ 0.36 &   {\color{red}{ 0.18}}    & *\\
    ResNeXt101,8x32   & SPA~\cite{guo2020spanet}   &  52.95M     &  81.27$\pm$ 0.31 &   {\color{red}{ 0.18}}    & * \\
    ResNeXt101,8x32   & SEM~\cite{zhong2022switchable}    & 34.09M      & 81.30$\pm$ 0.37 &   {\color{red}{ 0.21}}    & * \\
    ResNeXt101,8x32   & SPEM~\cite{zhong2022mix}   & 32.18M      &  81.29$\pm$ 0.40 &   {\color{red}{ 0.20}}    & * \\
    \rowcolor{Gray}ResNeXt101,8x32   & DIA-LSTM & 33.01M      &   \pmb{82.60$\pm$ 0.19} &   {\color{red}{ \pmb{1.51}}}    & N/A \\
    \rowcolor{Gray}ResNeXt101,8x32   & DIA-LSTM (Light) &  32.24M      &   82.37$\pm$ 0.20 &   {\color{red}{ 1.28}}    & * \\
    \bottomrule
    \end{tabular}%
    }
  \label{tab:cifar}%
\end{table}%

\section{Experiments}
\label{sec:exp}
In this section, \hzz{we evaluate the performance of the proposed DIA-LSTM and DIA-LSTM (Light) for enhancing different neural network backbones on multiple benchmark tasks.} \hzz{These tasks include image classification, image generation using the diffusion model, object detection, and a medical application. The chosen backbones consist of various ResNet variations, vision transformers and UNet.} 

In our experiments, we employ the two-sample Student's t-test as a statistical test to ensure that any differences observed between the accuracies of DIA-LSTM (or its lightweight variant) and other models are statistically significant. The significance level ($\alpha$) is set to 0.05. If the p-value is smaller than $\alpha$, it indicates a statistically significant performance distinction between the tested model and DIA-LSTM. Furthermore, if the accuracy of DIA-LSTM surpasses that of a self-attention module, it implies a significant improvement achieved by our DIA-LSTM.

\subsection{Image Classification}
\label{sec:cls}
We compare the proposed DIA-LSTM and DIA-LSTM (Light) with seven widely-used SAMs, including SE~\cite{hu2018squeeze}, ECA~\cite{wang2020eca}, CBAM~\cite{woo2018cbam}, SRM~\cite{lee2019srm}, SPA~\cite{guo2020spanet}, SEM~\cite{zhong2022switchable} and SPEM~\cite{zhong2022mix} across four image classification datasets. These datasets include CIFAR100~\cite{krizhevsky2009learning}, STL-10~\cite{pmlr-v15-coates11a}, miniImageNet~\cite{NIPS2016_90e13578} and ImageNet~\cite{ILSVRC15}. A summary of these datasets is presented in Table~\ref{tab:dataclassification}.

\begin{table}[htbp]
  \centering
  \caption{Summary of the datasets for image classification experiments.}
    \begin{tabular}{lrrrr}
    \toprule
    \textbf{Dataset} & \multicolumn{1}{l}{\textbf{\#class}} & \multicolumn{1}{l}{\textbf{\#training}} & \multicolumn{1}{l}{\textbf{\#testing}} & \textbf{Image size} \\
    \midrule
    CIFAR10 & 10    & 50,000 & 10,000 & 32 x 32 \\
    CIFAR100 & 100   & 50,000 & 10,000 & 32 x 32 \\
    STL-10 & 10    & 5,000 & 8,000 & 96 x 96 \\
    miniImageNet & 100   & 50,000 & 10,000 & 224 x 224 \\
    \hzz{ImageNet} & \hzz{1000}  & \hzz{1,281,123} & \hzz{50,000} & \hzz{224 x 224} \\
    \bottomrule
    \end{tabular}%
  \label{tab:dataclassification}%
\end{table}%

For CIFAR100 and STL-10 datasets, we utilize the well-known ResNet164~\cite{he2016deep}, WRN52-4~\cite{wrn}, and ResNeXt101,8$\times$32~\cite{xie2017aggregated} as the backbone models. The training settings for the experiments involving ResNet164 are identical to those specified in Table \ref{tab:diaformodule}. As for the other two backbone models, we adopt the settings outlined in \cite{huang2020dianet}.

\begin{table}[htbp]
  \centering
  \caption{Comparison of performance between existing self-attention modules and our DIA-LSTM on STL-10 dataset. The ``Org'' represents the backbone without any self-attention module. The p-value ($p$) indicates the results of the Student's t-test, comparing the accuracies of DIA-LSTM with other models. The significance level $(\alpha)$ is set at 0.05. Notation: `` * '': $p < 0.05$. The best and second-best results for each backbone are highlighted in bold and \underline{\textit{italic}} fonts, respectively.}
    \begin{tabular}{llcc}
    \toprule
    \textbf{Backbone} & \multicolumn{1}{c}{\textbf{Model}} & \multicolumn{1}{c}{\textbf{Top1-acc.}} & \multicolumn{1}{c}{\textbf{p-value}} \\
    \midrule
    ResNet164 & Org &   82.48$\pm$ 1.25    & * \\
    ResNet164 & SE~\cite{hu2018squeeze}    &  83.97$\pm$ 0.94     & * \\
    ResNet164 & ECA~\cite{wang2020eca}    &  83.76$\pm$ 1.34     & * \\
    ResNet164 & CBAM~\cite{woo2018cbam}  &  84.01$\pm$ 0.64     & * \\
    ResNet164 & SRM~\cite{lee2019srm}   &  84.32$\pm$ 0.94     & * \\
    ResNet164 & SPA~\cite{guo2020spanet}   &  83.99$\pm$ 0.98     & * \\
    ResNet164 & SPEM~\cite{zhong2022mix}   & 84.03$\pm$ 1.01     & * \\
    ResNet164 & SEM~\cite{zhong2022switchable}   &  85.09$\pm$ 0.72     & * \\
    \rowcolor{Gray}ResNet164 & DIA-LSTM &     \pmb{85.92$\pm$ 0.36}     & N/A\\
    \rowcolor{Gray}ResNet164 & DIA-LSTM (Light) &     \underline{\textit{85.80$\pm$ 0.29}}     & * \\
    \midrule
    WRN52-4 & Org & 87.42$\pm$ 1.38    & *  \\
    
    WRN52-4 & SE~\cite{hu2018squeeze}    & 88.25$\pm$ 1.68    & *  \\
    WRN52-4 & ECA~\cite{wang2020eca}    & 88.57$\pm$ 1.75    & *  \\
    WRN52-4 & CBAM~\cite{woo2018cbam}  & 87.38$\pm$ 2.06    & *  \\
    WRN52-4 & SRM~\cite{lee2019srm}   & 88.48$\pm$ 1.64    & * \\
    WRN52-4 & SPA~\cite{guo2020spanet}   & 88.32$\pm$ 1.65    & *\\
    WRN52-4 & SPEM~\cite{zhong2022mix}   & 87.68$\pm$ 2.25    & * \\
    WRN52-4 & SEM~\cite{zhong2022switchable}   &  88.19$\pm$ 1.46    & *\\
    \rowcolor{Gray}WRN52-4 & DIA-LSTM & \pmb{89.29$\pm$ 1.40}      & N/A \\
    \rowcolor{Gray}WRN52-4 & DIA-LSTM (Light) & \underline{\textit{89.00$\pm$ 1.62}}      & * \\
    \bottomrule
    \end{tabular}%
  \label{tab:stl10}%
\end{table}%

\begin{table}[htbp]
  \centering
  \caption{ Comparison of performance between existing self-attention modules and our DIA-LSTM on mini-ImageNet dataset. The ``Org'' represents the backbone without any self-attention module. The p-value ($p$) indicates the results of the Student's t-test, comparing the accuracies of DIA-LSTM with other models. The significance level $(\alpha)$ is set at 0.05. Notation: `` * '': $p < 0.05$. The best and second-best results for each backbone are highlighted in bold and \underline{\textit{italic}} fonts, respectively.}
    \begin{tabular}{llcc}
    \toprule
    \textbf{Backbone} & \multicolumn{1}{c}{\textbf{Model}} & \multicolumn{1}{c}{\textbf{Top1-acc.}} & \multicolumn{1}{c}{\textbf{p-value}} \\
    \midrule
    
    ResNet18 & Org &   77.18$\pm$ 0.07    & * \\
    ResNet18 & SE~\cite{hu2018squeeze}    & 77.98$\pm$ 0.10      & * \\
    ResNet18 & ECA~\cite{wang2020eca}    & 78.07$\pm$ 0.06      &  *\\
    ResNet18 & CBAM~\cite{woo2018cbam}  & 77.78$\pm$ 0.08 &  *\\
    ResNet18 & SRM~\cite{lee2019srm}   & 78.21$\pm$ 0.13 &  *\\
    ResNet18 & SPA~\cite{guo2020spanet}   & 77.99$\pm$ 0.07      & * \\
    \rowcolor{Gray}ResNet18 & DIA-LSTM &  \underline{\textit{78.88$\pm$ 0.06}}     & N/A \\
    \rowcolor{Gray}ResNet18 & DIA-LSTM (Light) & \pmb{79.07$\pm$ 0.10}      & * \\
    \midrule
    ResNet34 & Org &  78.04$\pm$ 0.10     & * \\
    ResNet34 & SE~\cite{hu2018squeeze}    &   \underline{\textit{78.88$\pm$ 0.09}}     & * \\
    ResNet34 & ECA~\cite{wang2020eca}    &   78.64$\pm$ 0.10    & * \\
    ResNet34 & CBAM~\cite{woo2018cbam}  &   78.44$\pm$ 0.07    & * \\
    ResNet34 & SRM~\cite{lee2019srm}   &   78.37$\pm$ 0.10    & * \\
    ResNet34 & SPA~\cite{guo2020spanet}   &   78.69$\pm$ 0.09    & * \\
    \rowcolor{Gray}ResNet34 & DIA-LSTM &  \pmb{79.84$\pm$ 0.09}     &  N/A\\
    \rowcolor{Gray}ResNet34 & DIA-LSTM (Light) &   78.83$\pm$ 0.10     &  *\\
    \midrule
    ResNet50 & Org &   79.39$\pm$ 0.12    &  * \\
    ResNet50 & SE~\cite{hu2018squeeze}    &  80.11$\pm$ 0.05    &  * \\
    
    ResNet50 & ECA~\cite{wang2020eca}    &  80.48$\pm$ 0.07    &  * \\
    ResNet50 & CBAM~\cite{woo2018cbam}  &  80.63$\pm$ 0.10    &  * \\
    ResNet50 & SRM~\cite{lee2019srm}   &  80.77$\pm$ 0.12    &  * \\
    ResNet50 & SPA~\cite{guo2020spanet}   &  80.26$\pm$ 0.15    &  * \\
    \rowcolor{Gray}ResNet50 & DIA-LSTM &  \pmb{81.37$\pm$ 0.10}    &  N/A \\
    \rowcolor{Gray}ResNet50 & DIA-LSTM (Light) &   \underline{\textit{81.00$\pm$ 0.07}}    &  * \\

    \bottomrule
    \end{tabular}%
  \label{tab:miniimagenet}%
\end{table}%

\begin{table}[htbp]
  \centering
  \caption{
  \hzz{Comparison of performance between existing self-attention modules and our DIA-LSTM on ImageNet dataset. The ``Org'' represents the backbone without any self-attention module. The p-value ($p$) indicates the results of the Student's t-test, comparing the accuracies of DIA-LSTM with other models. The significance level $(\alpha)$ is set at 0.05. Notation: `` * '': $p < 0.05$. The best and second-best results for each backbone are highlighted in bold and \underline{\textit{italic}} fonts, respectively.}}
    \begin{tabular}{llrc}
    \toprule
    \textbf{Backbone} & \multicolumn{1}{c}{\textbf{Model}} & \multicolumn{1}{c}{\textbf{Top1-acc.}} & \multicolumn{1}{c}{\textbf{p-value}}\\

    \midrule
    ResNet18 & Org &   69.98$\pm$ 0.08  & *    \\
    ResNet18 & SE~\cite{hu2018squeeze}    & 70.57$\pm$ 0.04   & *     \\
    ResNet18 & ECA~\cite{wang2020eca}    & 70.36$\pm$ 0.10   & *     \\
    ResNet18 & CBAM~\cite{woo2018cbam}  & 70.77$\pm$ 0.03    & *    \\
    ResNet18 & SRM~\cite{lee2019srm}   & 70.60$\pm$ 0.07     & *   \\
    ResNet18 & SPA~\cite{guo2020spanet}   & 70.82$\pm$ 0.11   & *     \\
    \rowcolor{Gray}ResNet18 & DIA-LSTM &   \pmb{71.30$\pm$ 0.16}   & N/A   \\
    \rowcolor{Gray}ResNet18 & DIA-LSTM (Light) &   \underline{\textit{70.97$\pm$ 0.09}}   & *   \\
    \midrule
    ResNet34 & Org &  73.99$\pm$ 0.04 & *\\
    ResNet34 & SE~\cite{hu2018squeeze}    & 74.37$\pm$ 0.10 & *\\
    ResNet34 & ECA~\cite{wang2020eca}    & 74.56$\pm$ 0.08 & *\\
    ResNet34 & CBAM~\cite{woo2018cbam}  & \underline{\textit{74.89$\pm$ 0.11}} & *\\
    ResNet34 & SRM~\cite{lee2019srm}   &  74.79$\pm$ 0.04& *\\
    ResNet34 & SPA~\cite{guo2020spanet}   & 74.59$\pm$ 0.14 & *\\
    \rowcolor{Gray}ResNet34 & DIA-LSTM &  \pmb{75.09$\pm$ 0.03}& N/A\\
    \rowcolor{Gray}ResNet34 & DIA-LSTM (Light) & 74.79$\pm$ 0.09 & *\\
    \midrule
    ResNet50 & Org & 76.02$\pm$ 0.09 \\
    ResNet50 & SE~\cite{hu2018squeeze}    & 76.62$\pm$ 0.02 & *\\
    ResNet50 & ECA~\cite{wang2020eca}    & 77.07$\pm$ 0.09 & *\\
    ResNet50 & CBAM~\cite{woo2018cbam}  & 76.38$\pm$ 0.10 & *\\
    ResNet50 & SRM~\cite{lee2019srm}   & 76.49$\pm$ 0.03 & *\\
    ResNet50 & SPA~\cite{guo2020spanet}   &  77.02$\pm$ 0.12& *\\
    \rowcolor{Gray}ResNet50 & DIA-LSTM & \pmb{77.52$\pm$ 0.06} & N/A\\
    \rowcolor{Gray}ResNet50 & DIA-LSTM (Light) &  \underline{\textit{77.12$\pm$ 0.04}}  & *\\

    \bottomrule
    \end{tabular}%
  \label{tab:imagenet}%
\end{table}%

\begin{table}[htbp]
  \centering
  \caption{\hzz{The performance of vision transformer on ImageNet~\cite{dosovitskiy2021an}. ``Org" denotes the original backbone.  ``ViT-B/16" denote that we adopt the base model of vision transformer with patch size 16.}}
  \resizebox*{0.9\linewidth}{!}{
    \begin{tabular}{llcc}
    \toprule
    \multicolumn{1}{c}{\textbf{Backbone}} & \multicolumn{1}{c}{\textbf{Model}} & \textbf{Top1-acc.} & \textbf{p-value} \\
    \midrule
    ViT-S/16 & Org   & 80.05$\pm$ 0.10 & * \\
    ViT-S/16 & SE~\cite{hu2018squeeze}   & 80.38$\pm$ 0.08 & * \\
    ViT-S/16 & ECA~\cite{wang2020eca}   & 80.46$\pm$ 0.12 & * \\
    ViT-S/16 & SPA~\cite{guo2020spanet}   & 80.29$\pm$ 0.08 & * \\
    \rowcolor{Gray}ViT-S/16 & DIA-LSTM & \pmb{80.95}$\pm$ \pmb{0.07} & N/A \\
    \rowcolor{Gray}ViT-S/16 & DIA-LSTM (Light) & 80.86$\pm$ 0.09 & * \\
    \midrule
    ViT-B/16 & Org   & 81.39$\pm$ 0.12 & * \\
    ViT-B/16 & SE~\cite{hu2018squeeze}   & 82.48$\pm$ 0.14 & * \\
    ViT-B/16 & ECA~\cite{wang2020eca}   & 82.18$\pm$ 0.09 & * \\
    ViT-B/16 & SPA~\cite{guo2020spanet}   & 82.24$\pm$ 0.07 & * \\
    \rowcolor{Gray}ViT-B/16 & DIA-LSTM & \pmb{82.84}$\pm$ \pmb{0.07} & N/A \\
    \rowcolor{Gray}ViT-B/16 & DIA-LSTM (Light) & 82.76$\pm$ 0.10 & * \\
    \bottomrule
    \end{tabular}%
    }
  \label{tab:transf}%
\end{table}%

\textbf{CIFAR100.} Table \ref{tab:cifar} shows the classification results on CIFAR100. DIA-LSTM achieves a remarkable improvement in testing accuracy compared with the original networks across various datasets and backbones. Also, DIA-LSTM consistently outperforms other existing self-attention modules. Specifically, 
DIA-LSTM demonstrates impressive test accuracies of 77.09\%, 81.02\%, and 82.60\% on ResNet164, WRN52-4, and ResNeXt101,8$\times$32 backbones, respectively. These results position DIA-LSTM as either the best or second-best performing model among all the models listed in Table \ref{tab:cifar}. Considering that all the p-values in Table \ref{tab:cifar} are below 0.05, it confirms that the improvement results obtained by DIA-LSTM are statistically significant. Besides, the lightweight version of DIA-LSTM, known as DIA-LSTM (Light), achieves competitive results compared to DIA-LSTM while having a significantly smaller number of parameters.

\begin{table*}[htbp]
  \centering
  \caption{The object detection performance of existing self-attention modules and DIA-LSTM on MS COCO 2017. The p-value ($p$) is the Student’s t-test between the AP of the DIA-LSTM and other models. The
significance level $\alpha$ is 0.05. Notation: “*”: $p < 0.05$. The best and the second best results of each setting are marked in bold and \underline{\textit{italic}} fonts, respectively. “AP”, “AP$_{S}$”,
 “AP$_{M}$”, and “AP$_{L}$”: averaged AP for overall, small, medium, and large scale objects, respectively, at [50\%, 95\%] IoU interval with a step as 5\%, “AP$_{50}$” and “AP$_{75}$”: AP at IoU as 50\% and 75\%, respectively. The accuracy improvement is highlighted in {\color{red}{red}}.}
    \begin{tabular}{rlcccccccc}
    \toprule
    \multicolumn{1}{l}{\textbf{Detector}} & \textbf{Model} & \multicolumn{1}{l}{\textbf{AP }} & \multicolumn{1}{l}{\textbf{AP$_{50}$}} & \multicolumn{1}{l}{\textbf{AP$_{75}$}} & \multicolumn{1}{l}{\textbf{AP$_{S}$}} & \multicolumn{1}{l}{\textbf{AP$_{M}$}} & \multicolumn{1}{l}{\textbf{AP$_{L}$}} & \multicolumn{1}{l}{\textbf{↑AP}} & \multicolumn{1}{l}{\textbf{p-value}}\\
        \midrule
    \multicolumn{1}{l}{Faster R-CNN} & ResNet50 &   36.5   &    57.9   &   39.2    &    21.5   &   39.9  &    46.6   & - & *\\

          & +SE~\cite{hu2018squeeze}    &  38.0     &  60.3     & 41.2      &  22.9     &  41.8     &    49.0   & {\color{red}{ 1.5}} & *\\
          & +ECA~\cite{wang2020eca}    &  38.4     &  \pmb{60.6}     &  41.0     &  23.3     &  \underline{\textit{42.7}}     &  48.5     & {\color{red}{ 1.9}} & *\\
          & +SGE~\cite{li2019spatial}   &  38.5     &  \underline{\textit{60.5}}     &  41.7     &  \underline{\textit{23.5}}     &  42.5     &  49.1     & {\color{red}{ 2.0}} & *\\
          \rowcolor{Gray}& +DIA-LSTM & \pmb{39.1}      &  60.2     &  \pmb{42.4}     &  \pmb{23.9}     & \pmb{42.9}      & \pmb{50.0}      & {\color{red}{ \pmb{2.6}}} & N/A\\
          \rowcolor{Gray}& +DIA-LSTM (Light) & \underline{\textit{38.7}}     &   60.2    &  \underline{\textit{41.8}}     &  23.3     &  42.5     &  \underline{\textit{49.7}}     &{\color{red}{\underline{\textit{2.2}}}}  &*\\
          \midrule
    \multicolumn{1}{l}{Faster R-CNN} & ResNet101 & 38.6      & 60.1      & 41.0      & 22.0      & 43.6      & 49.8      & - &*\\
          & +SE~\cite{hu2018squeeze}    &  39.8     &    61.7   &    43.8   &    23.3   &    44.3   &    51.3   & {\color{red}{ 1.2}} & *\\
          & +ECA~\cite{wang2020eca}    &  40.1     &    \underline{\textit{62.5}}   &    \underline{\textit{44.0}}   &    24.0   &    44.6   &    51.2   & {\color{red}{ 1.5}}  &* \\
          & +SGE~\cite{li2019spatial}   &  40.3     &    61.7   &    43.9   &    24.3   &    44.9   &    51.0   & {\color{red}{ 1.7}} & *\\
          \rowcolor{Gray}& +DIA-LSTM &  \underline{\textit{41.3}}     & \pmb{62.7}      & \pmb{44.3}      &  \pmb{24.5}      & \pmb{46.5}      &  \underline{\textit{52.9}}      & {\color{red}{  \underline{\textit{2.7}}}} & N/A\\
          \rowcolor{Gray}& +DIA-LSTM (Light) &  \pmb{41.8}    &   62.0    & 43.7      & \pmb{24.5}      & \underline{\textit{46.3}}      & \pmb{53.8}      & {\color{red}{ \pmb{3.1}}} &*\\
          \midrule
    \multicolumn{1}{l}{RetinaNet} & ResNet50 &  35.5     & 56.4      & 38.5      & 20.3      & 39.7      & 46.0      & - & *\\
          & +SE~\cite{hu2018squeeze}    &  37.4     &    57.5   &    39.6   &    21.5   &    40.5   &    49.6   & {\color{red}{ 1.9}} &*\\
          & +ECA~\cite{wang2020eca}    &  37.4     &    57.3   &    39.6   &    21.4   &    \underline{\textit{41.5}}   &    48.7   & {\color{red}{ 1.9}} &*\\
          & +SGE~\cite{li2019spatial}   &  37.6     &    57.8   &    \underline{\textit{39.7}}   &    22.0   &    41.0   &    49.1   & {\color{red}{ 2.1}} &*\\
          \rowcolor{Gray}& +DIA-LSTM &  \pmb{38.8}     & \pmb{58.0}      & \pmb{40.3}      & \pmb{23.4}      & \pmb{42.4}      & \pmb{50.1}      & {\color{red}{ \pmb{3.3}}} & N/A\\
          \rowcolor{Gray}& +DIA-LSTM (Light) &  \underline{\textit{37.9}}      & \underline{\textit{57.9}}      & 39.6      & \underline{\textit{22.6}}      & 41.3      & \underline{\textit{49.7}}      & {\color{red}{ \underline{\textit{2.4}}}} & *\\
    \midrule
    \multicolumn{1}{l}{RetinaNet} & ResNet101 &  37.8     & 57.8       &  39.8     &  21.3     &  42.5     & 49.5      &- & *\\
          & +SE~\cite{hu2018squeeze}    &  38.6     & 59.3       &  41.4     &   22.1    &  43.2     &  50.2     & {\color{red}{ 0.8}} & *\\
          & +ECA~\cite{wang2020eca}    &  39.4     &  \pmb{59.7}     &  41.6     &  22.9     &  \underline{\textit{43.8}}     & 51.1      &{\color{red}{ 1.6}}  &*\\
          & +SGE~\cite{li2019spatial}   &  38.8     &  59.0     &   41.4    &   22.6    &   43.2    &   50.2    & {\color{red}{ 1.0}} &*\\
          \rowcolor{Gray}& +DIA-LSTM &  \pmb{40.2}     &  \underline{\textit{59.6}}     & \pmb{42.8}      &  \pmb{24.3}     &  \pmb{44.1}     & \underline{\textit{51.8}}      & {\color{red}{ \pmb{2.4}}} & N/A\\
          \rowcolor{Gray}& +DIA-LSTM (Light) & \underline{\textit{39.8}}      &   59.4    &   \underline{\textit{42.0}}    &  \underline{\textit{23.6}}     &  43.6     &   \pmb{52.0}    & {\color{red}{ \underline{\textit{2.0}}}} &*\\
    \bottomrule
    \end{tabular}%
  \label{tab:object}%
\end{table*}%

\textbf{STL-10.} Table \ref{tab:stl10} presents the classification results on the STL-10 dataset. Once again, DIA-LSTM emerges as the top-performing module across the ResNet164 and WRN52-4 backbones, achieving impressive classification accuracies of 85.92\% and 89.29\%, respectively. Given the higher resolution of the images, the classification task on the STL-10 dataset is known to be more challenging compared to CIFAR10/100. Yet, the results shown in Table \ref{tab:stl10} demonstrate that DIA-LSTM can achieve commendable performance on more challenging tasks. This further underscores the effectiveness of our proposed method. 

\textbf{Imagenet and mini-Imagenet.} We present the results of mini-Imagenet \hzz{and ImageNet} in Table \ref{tab:miniimagenet} and Table \ref{tab:imagenet}, respectively. DIA-LSTM consistently achieves the best or second-best performance across ResNet models with different depths. When compared to the corresponding "Org" models, the models incorporating the proposed DIA-LSTM demonstrate a substantial improvement in performance. Besides, DIA-LSTM exhibits statistically significant performance improvements on the mini-ImageNet and ImageNet datasets, as evidenced by the p-values in both Table \ref{tab:miniimagenet} and Table \ref{tab:imagenet} being smaller than 0.05. 

\hzz{Previous works~\cite{zhong2023asr,wang2023convolution,wang2021evolving} have demonstrated that additional attention modules can enhance the performance of the Vision Transformer (ViT)~\cite{dosovitskiy2021an}. Motivated by their findings and considering the experimental settings in \cite{zhong2023asr}, we explore the application of the DIA unit to the ViT architecture. Although these improvements may not necessarily achieve state-of-the-art performance, the results presented in Table \ref{tab:transf} consistently demonstrate that DIA enhances the performance of the ViT models.}

\subsection{Image Generation}
\begin{figure}[t]
	 		\centering
	 		\includegraphics[width=0.8\linewidth]{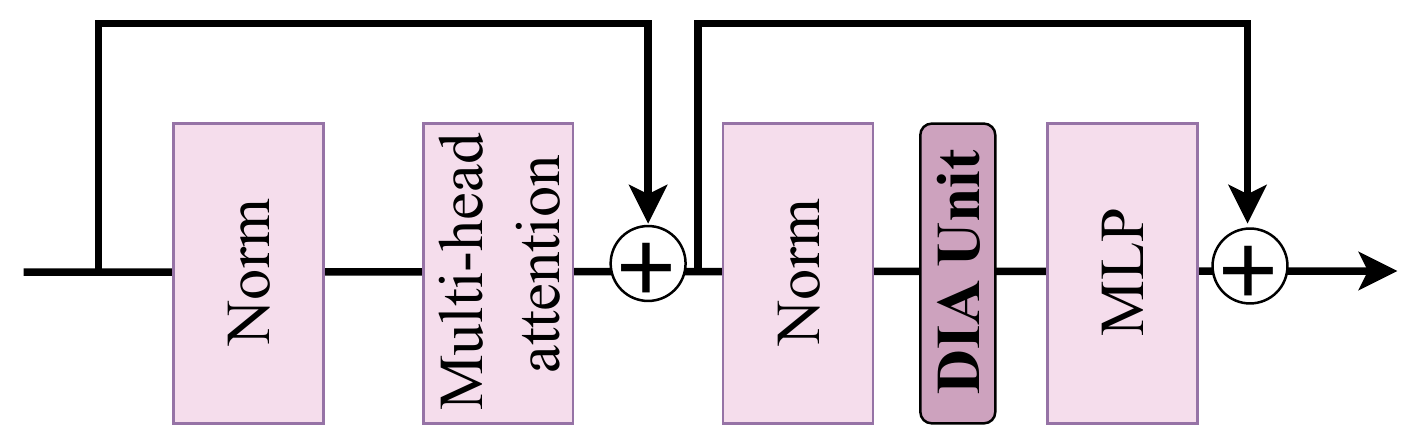}
	 		\caption{\hzz{Equipping UViT with the DIA unit. Motivated by the setting in \cite{zhong2023asr}, we place the DIA unit between the normalization~(Norm) layer and the multi-layer perception (MLP) layer.}} 
	 \vspace{-0.2cm}
	 		\label{fig:transformer}
	 	\end{figure}

\begin{figure}[t]
	 		\centering
	 		\includegraphics[width=0.9\linewidth]{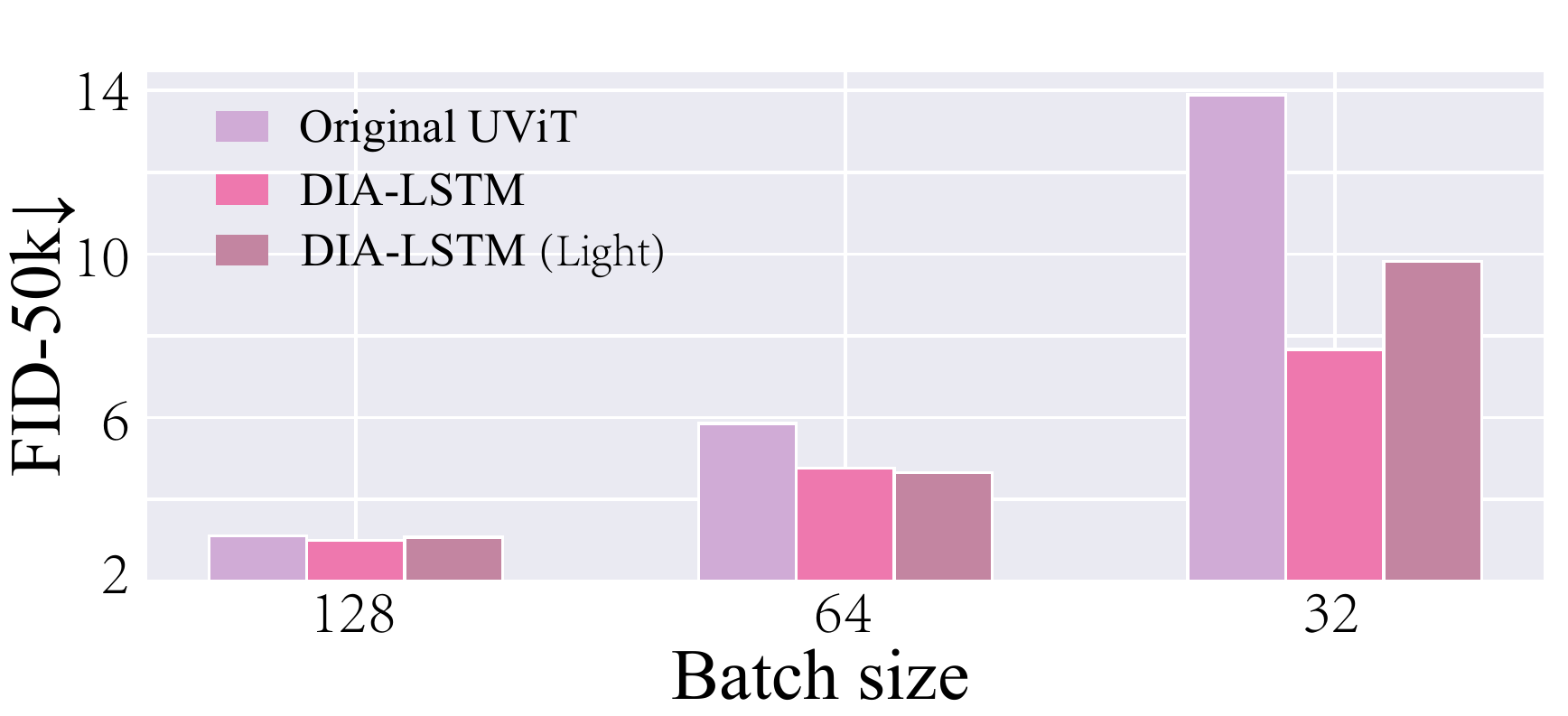}
	 		\caption{\hzz{Synthesis performance of proposed methods on CIFAR10. UViT~\cite{bao2023all} is a popular backbone network for the diffusion model. To evaluate the quality of the generated images, we utilize the Fréchet Inception Distance (FID) as a metric. The FID measures the dissimilarity between the distribution of generated images and a reference set of real images. A smaller FID score indicates a better result.}} 
	 \vspace{-0.2cm}
	 		\label{fig:diffusion}
	 	\end{figure}

\hzz{Recently,  diffusion models~\cite{lugmayr2022repaint,hang2023efficient,ho2020denoising,zhong2023adapter} have become one of the most successful generative models because of their superiority in  modeling realistic data distributions. In this section, we consider a UNet-based backbone network UViT~\cite{bao2023all}, which is a novel diffusion model with a Transformer structure, and investigate the performance of the proposed DIA in image generation tasks. 

Specifically, we apply DIA to UViT following the architecture depicted in Fig.\ref{fig:transformer}. The results in Fig.\ref{fig:diffusion} demonstrate that the introduced DIA can indeed enhance the quality of generated images, particularly when the batch size is small, such as 32. In practice, UViT is often sensitive to batch sizes~\cite{bao2023all} as smaller batches tend to introduce more noise in the stochastic gradients, further exacerbating gradient instability. Therefore, the empirical results in Fig.~\ref{fig:diffusion} empirically indicate that DIA can contribute to the stable training of the model. For a more detailed and comprehensive discussion on the relationship between DIA and stable training, please refer to Section \ref{sec:stab}.}

\subsection{Object Detection}

In light of the significant improvements observed in classification tasks in Section \ref{sec:cls}, this section delves into exploring the performance of DIA-LSTM for object detection tasks. The experiments are carried out on the MS COCO dataset. Specifically, we compare the performance of DIA-LSTM and DIA-LSTM (Light) with three popular self-attention modules, namely SE~\cite{hu2018squeeze}, ECA~\cite{wang2020eca}, and SGE~\cite{li2019spatial}. These comparisons are conducted using the same settings as~\cite{zhao2021recurrence}. We utilize the Faster R-CNN and RetinaNet detectors implemented through the MMDetection toolkit, an open-source framework. The MS COCO dataset contains 80 classes with 118,287 training images and 40,670 test images. The results are presented in Table \ref{tab:object}. Each model is verified 15 times with random seeds to ensure robustness and reliability. The reported metrics include p-value, Average Precision (AP), and other standard COCO evaluation metrics.

For the Faster R-CNN detector, we observe that our DIA-LSTM model achieves superior performance compared to the baseline models. Specifically, DIA-LSTM outperforms the baseline models by 2.6\% and 2.7\% in terms of AP on ResNet50 and ResNet101, respectively.

Similarly, for the RetinaNet detector, incorporating DIA-LSTM into the ResNet backbone consistently yields exceptional results, surpassing all other models by a significant margin. Furthermore, the p-values obtained from statistical tests support the significant improvement achieved by DIA-LSTM in terms of performance enhancement.

	 	\begin{figure}[t]
	 		\centering
	 		\includegraphics[width=0.99\linewidth]{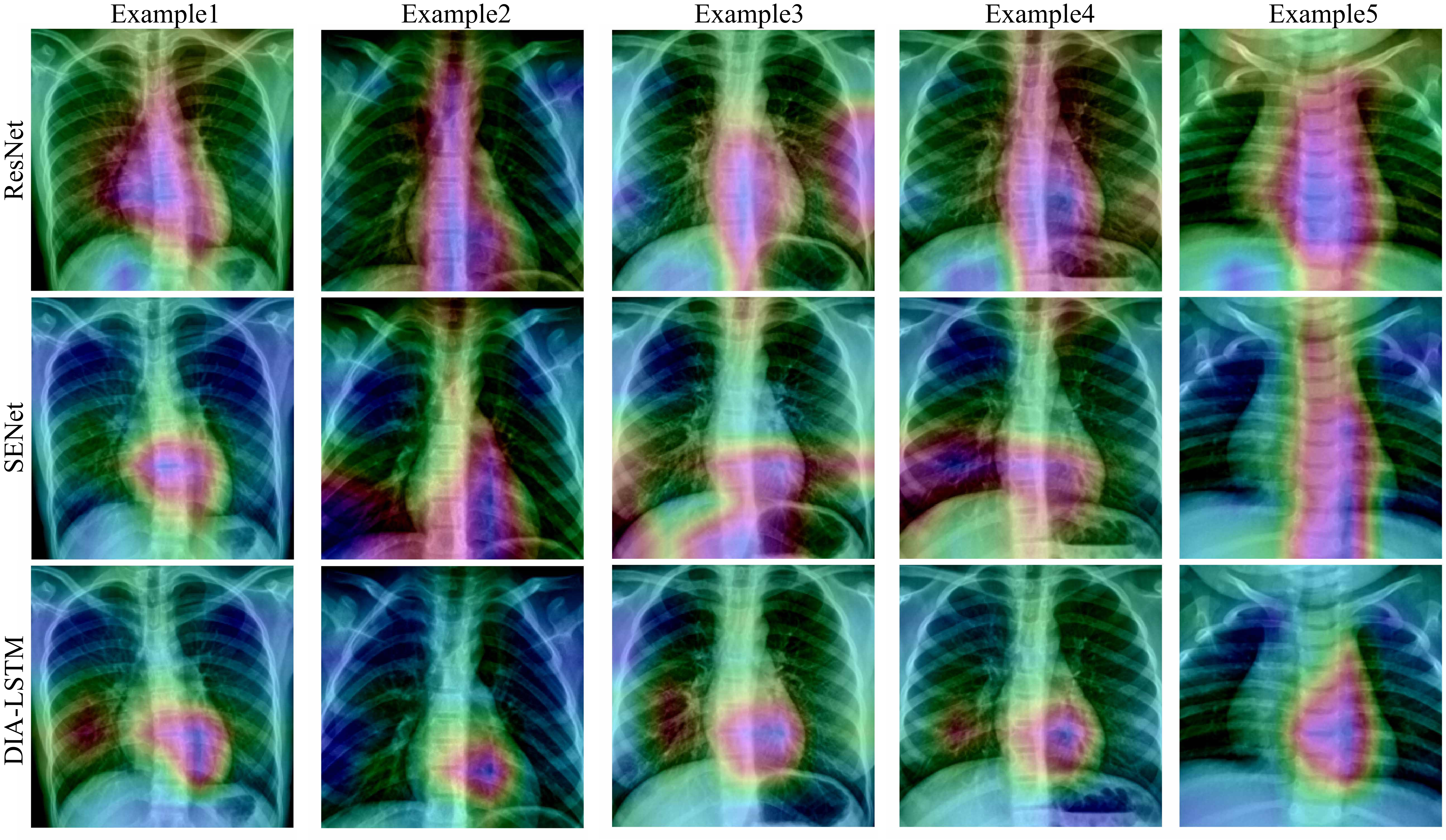}
	 		\caption{Grad-CAM visualizations for different models. The purple regions indicate critical areas that contribute significantly to the network's output of classifying as  Pneumonia, while the dark regions represent less relevant areas. } 
	 \vspace{-0.2cm}
	 		\label{fig:medicalcam}
	 	\end{figure}
\subsection{Medical Application}
We employ the proposed DIA-LSTM method in a medical application, specifically the classification of chest X-ray images for pneumonia. The primary objective is to showcase the ability of our approach to enhance accuracy and capture discriminative features. We adopt a dataset obtained from~\cite{kermany2018identifying}, which comprises 5,863 X-ray images classified into two categories: Pneumonia and Normal. 

\begin{table}[htbp]
  \centering
  \caption{The accuracy of chest x-ray image classification for pneumonia.}
    \begin{tabular}{lcccc}
    \toprule
    \textbf{Model} & \multicolumn{1}{l}{\textbf{Pneumonia}} & \multicolumn{1}{l}{\textbf{Normal}} & \multicolumn{1}{l}{\textbf{All}} & \multicolumn{1}{l}{\textbf{p-value}} \\
    \midrule
    ResNet50 &   97.8    &  52.5     &   81.4    &  *\\
    +SE~\cite{hu2018squeeze}    &   99.2    &  60.8     &  85.3     & * \\
    \rowcolor{Gray}+DIA-LSTM &   \pmb{99.4}    &  \pmb{71.1}     &  \pmb{89.2}     & N/A \\
    \rowcolor{Gray}+DIA-LSTM (Light) &   \pmb{99.4}    &  63.7     &  86.6     & * \\
    \bottomrule
    \end{tabular}%
  \label{tab:pneu}%
\end{table}%

\textbf{Classification accuracy.} The experimental outcomes for the chest X-ray image classification task, including the mean classification accuracies over 15 repeated experiments, are presented in Table \ref{tab:pneu}. Note that all models in Table \ref{tab:pneu} are trained from scratch, without utilizing any pre-trained models. In Table \ref{tab:pneu}, we observe that due to the imbalanced number of data in each class, all models are capable of classifying the Pneumonia class accurately, whereas they encounter challenges in correctly distinguishing normal X-ray images. Therefore, the main difficulty of this task lies in accurately identifying the normal class. The proposed DIA-LSTM exhibits the best accuracy among the models, while DIA-LSTM (Light) achieves competitive results compared to SE yet possesses significantly fewer parameters.

\textbf{Capturing discriminative features.} To evaluate the ability of DIA-LSTM in capturing and leveraging features relevant to a given target, which is crucial for explainable medical applications, we utilize Grad-CAM (Gradient-weighted Class Activation Mapping)~\cite{selvaraju2017grad}. Grad-CAM is a well-known tool used to generate heatmaps that highlight network attention based on gradients related to the target being analyzed. In Figure \ref{fig:medicalcam}, we present the visualization results and softmax scores for the target using the original ResNet50, SENet, and DIA-LSTM models on the X-ray dataset. In the visualization, the purple region indicates an important area that contributes significantly to the network's output, while the dark region represents areas that are less relevant.

The obtained results demonstrate that DIA-LSTM is capable of extracting similar features as SENet, while also capturing more intricate details of the target. Specifically, the regions of focus for DIA-LSTM are noticeably concentrated in the lungs, indicating higher confidence in its predictions. On the other hand, the original ResNet tends to prioritize the region in the middle of the image, namely the spine, rather than the lungs. These findings reveal that DIA-LSTM potentially possesses a more crucial ability to emphasize discriminative features for distinguishing between normal and pneumonia classes compared to both the original ResNet and SENet.

	\begin{table}[t]
	    \centering
			\caption{Test accuracy (\%) with different reduction ratio on CIFAR100 with ResNet164. The value in the bracket means the parameter increment compared with the original ResNet164 (1.73M).}
			\centering
			\begin{tabular}{ccc}
				\midrule
				Reduction ratio $r$ & $\#$Param (M) & top1-acc. \\
				\midrule
				1  & 2.59$_{(+0.86)}$ & 77.24  \\
				4  & 1.95$_{(+0.22)}$  & 77.09  \\
				8  & 1.84$_{(+0.11)}$  & 76.68  \\
				16  & 1.78$_{(+0.05)}$  & 76.59  \\
				\bottomrule
			\end{tabular}%
			\label{tab:reduction ratio}%
	\end{table}{}

\section{Abaltion study}
\label{sec:ablation}

In this section, we conduct ablation experiments to explore how to better plug DIA in different neural network structures and gain a deeper understanding of the role of components in the unit. All experiments are performed on CIFAR100 with ResNet for image classification. 

\subsection{\hzz{Different Numbers of SAM-incorporated Blocks}}
\begin{figure}[t]
	 		\centering
	 		\includegraphics[width=0.9\linewidth]{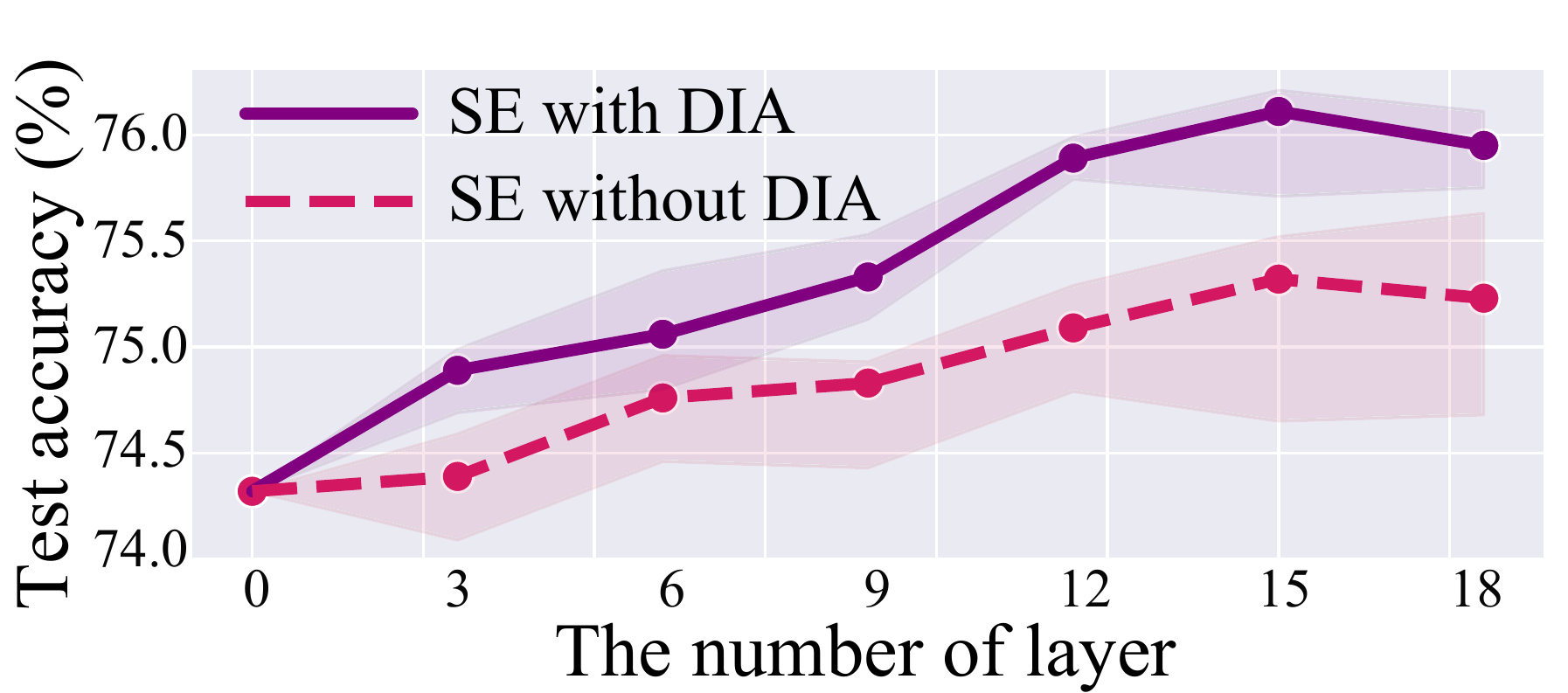}
	 		\caption{\hzz{The test accuracy with varied numbers of network layers with and without DIA. ``SE" is the attention module in SENet~\cite{hu2018squeeze}.} } 
	 \vspace{-0.2cm}
	 		\label{fig:numlayer}
	 	\end{figure}

\hzz{We explore how performance is affected by incorporating SAMs with or without DIA into different numbers of blocks. In this part, we utilize a ResNet164 model consisting of 18 blocks and select a subset of these blocks to incorporate SAMs. Figure~\ref{fig:numlayer} illustrates the testing accuracy achieved with varying numbers of blocks equipped with SAMs out of the total 18 blocks. We observe that the performance generally improves as more blocks are equipped with SAMs. However, the enhancement is more consistent and significant when SAMs are combined with DIA. These results indicate that the sharing mechanism can potentially amplify the benefits of SAM utilization.}

\subsection{Reduction Ratio in DIA-LSTM}
The reduction ratio $r$ is the sole hyperparameter in DIA-LSTM, employed to reduce the number of parameters utilized in LSTM. The main advantage of our model is improving the versatility of the DNN with a light parameter increment. The smaller reduction ratio causes a higher parameter increment and model complexity. This part investigates the trade-off between model complexity and performance. 

Table~\ref{tab:reduction ratio} presents the test accuracy of DIA-LSTM with different $r$. We observe that the number of parameters of the DIA-LSTM decreases with the increasing reduction ratio, but the testing accuracy declines slightly, which suggests that the model performance is not sensitive to the reduction ratio. In the case of $r=16$, the ResNet164 with DIA-LSTM has 0.05M parameter increment compared to the original ResNet164 but the testing accuracy of ResNet164 with DIA-LSTM is 76.59\% while that of the ResNet164 is 74.32\%. 

\subsection{Depth of Backbone with DIA-LSTM}
In practice, simply increasing the number of parameters in deep neural networks (DNNs) does not necessarily lead to significant performance improvements. Deeper networks can often suffer from excessive feature and parameter redundancy \cite{huang2017densely}. Therefore, designing a new structure of DNN is necessary~\cite{he2016deep,huang2017densely,srivastava2015training,hu2018squeeze,hu2018gather,wang2018non}. Since DIA-LSTM affects the topology of DNN backbones, evaluating the effectiveness of DIA-LSTM structure is important. Here we investigate how the depth of DNNs influences DIA-LSTM in two aspects: (1) the performance of DIA-LSTM compared to SENet of various depths; (2) the parameter increment of the ResNet with DIA-LSTM. 
		\begin{table}[ht]
			\centering
			\caption{Test accuracy (\%) with ResNet of different depth on CIFAR100.}
			\begin{tabular}{lcccc}
				\toprule
				\multicolumn{1}{c}{} & \multicolumn{2}{c}{SENet} & \multicolumn{2}{c}{DIA-LTSM ($r=4$)} \\
				\cmidrule{2-5}    \multicolumn{1}{c}{Depth} & $\#$P(M) & acc.  & $\#$P(M) & acc. \\
				\midrule
				ResNet83 & 0.99  & 74.68  & 1.11$_{(+0.12)}$ & 75.72  \\
				ResNet164 & 1.93  & 75.23  & 1.95$_{(+0.02)}$ & 77.09  \\
				ResNet245 & 2.87  & 75.43  & 2.78$_{(-0.09)}$ & 77.59  \\
				ResNet407 & 4.74  & 75.74  & 4.45$_{(-0.29)}$ & 78.08  \\
				\bottomrule
			\end{tabular}%

			\label{tab:DIANet with different depth1}%
	\end{table}{}	
	
	Table \ref{tab:DIANet with different depth1} presents the performance of SE and DIA-LSTM used in the ResNet backbones of varied depth. We observe that as the depth increases from 83 to 407 layers, the DIA-LSTM with a smaller number of parameters can achieve higher classification accuracy than the SENet. Moreover, the ResNet83 with DIA-LSTM can achieve a similar performance as the SENet407, and ResNet164 with DIA-LSTM outperforms all the SENets with at least 1.35\% and at most 58.8\% parameter reduction. They imply that the ResNet with DIA-LSTM is of higher parameter efficiency than SENet. The results also suggest that: as shown in Fig.~\ref{fig:diaunit}, the DIA-LSTM will pass more layers recurrently with a deeper depth. The DIA-LSTM can handle the interrelationship between the information of different layers in a much deeper DNN and figure out the long-distance dependency between layers. 

\subsection{Activation and Stacking Cells in DIA-LSTM}
We choose different activation functions in the output layer of DIA-LSTM in Fig.~\ref{fig:lstm}~(Middle) and different numbers of stacking LSTM cells to explore the effects of these two factors. 
    In Table \ref{tab:activation and number of unit}, we find that the performance has been significantly improved after replacing tanh in the standard LSTM with the sigmoid. As shown in Fig.~\ref{fig:lstm} (Middle), this activation function is located in the output layer and directly changes the effect of memory unit $c_t$ on the output of the output gate. In fact, the sigmoid is used in many attention-based methods like SENet as a gate mechanism. The test accuracy of different choices of LSTM activation functions in Table~\ref{tab:activation and number of unit} shows that sigmoid better help LSTM as a gate to rescale channel features. Table~12 in the SENet paper~\cite{hu2018squeeze} shows the performance of different activation functions like sigmoid $>$ tanh $>$ ReLU~(bigger is better), which coincides with our reported results and existing observation~\cite{liang2020instance}.
	
	When we use sigmoid in the output layer of LSTM, the increasing number of stacking LSTM cells does not necessarily lead to performance improvement but may lead to performance degradation. However, when we choose tanh, the situation is different. It suggests that, through the stacking of LSTM cells, the scale of the information flow among them is changed, which may affect the performance.

	\begin{table}[t]
		\centering
		\caption{Test accuracy (\%) on CIFAR100 with DIANet164 of different activation functions at the output layer in the modified LSTM and the different number of stacking LSTM cells.}
		
		\begin{tabular}{cccc}
			\toprule
			$\#$P(M) & Activation & $\#$LSTM cells& top1-acc. \\
			\midrule
			 1.95  & sigmoid & 1  & 77.09  \\
			 1.95  & tanh  & 1  & 75.64  \\
			 1.95  & ReLU & 1  & 75.02  \\
			 3.33  & sigmoid & 3  & 76.07  \\
			 3.33  & tanh  & 3  & 76.87  \\
			\bottomrule
		\end{tabular}

		\label{tab:activation and number of unit}
	\end{table}	


	 	\begin{figure*}[t]
	 		\centering
	 		\includegraphics[width=0.9\textwidth]{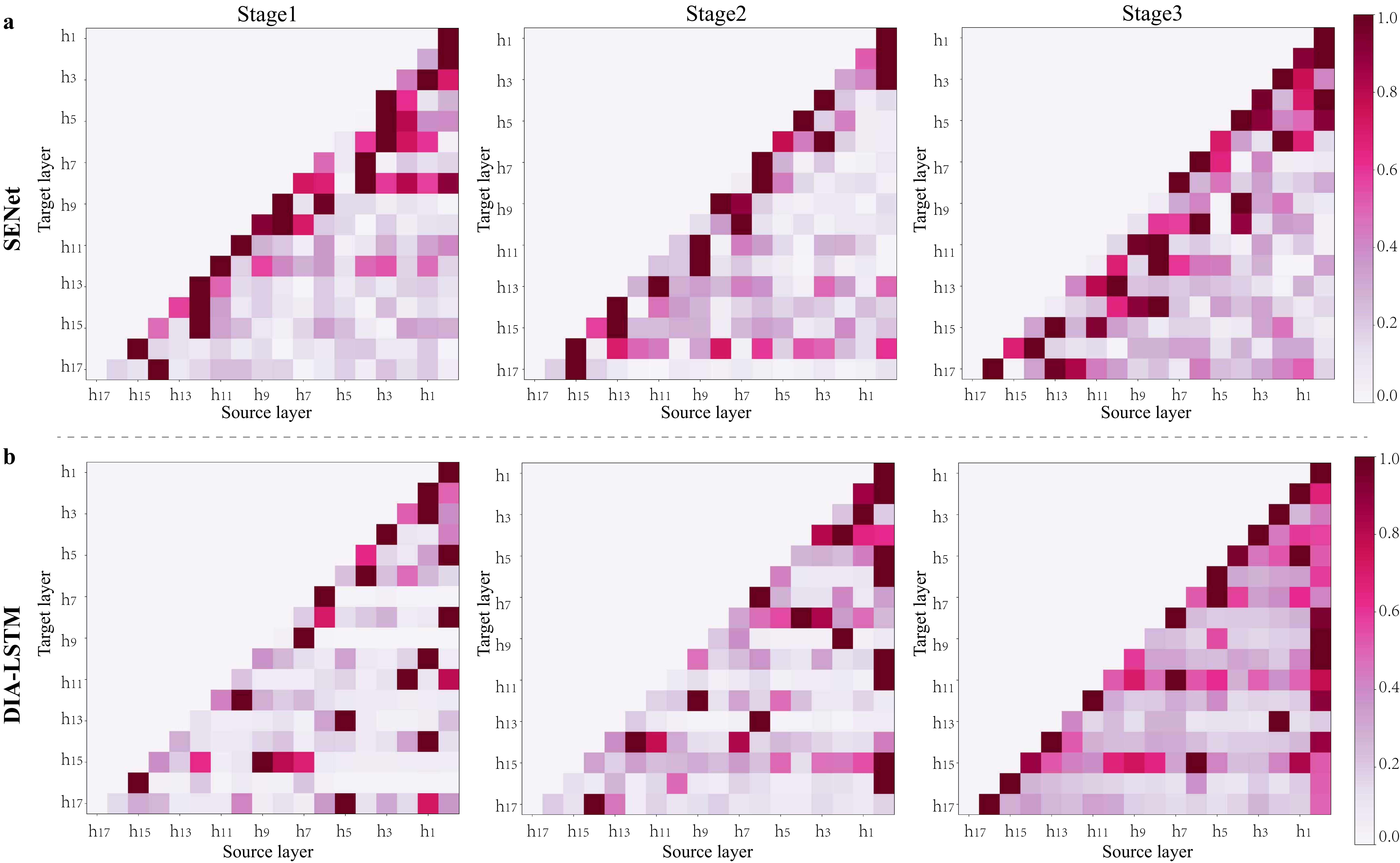}
	 				\vspace{-0.3cm}
	 		\caption{Visualization of feature integration for each stage by random forest. Each row presents the importance of source layers $h_n,1\leq n < t$ contributing to the target layer $h_t$.} 
	 \vspace{-0.2cm}
	 		\label{features integration for each stage}
	 	\end{figure*}

\section{Analysis}
\label{sec:analysis}

\subsection{Understanding Effectiveness of DIA}

In this section, we conduct several experiments for DIA-LSTM and analyze how DIA benefits the training from two views: (1) The effect on stabilizing training and (2) the dense-and-implicit connection.

\subsubsection{Dense-and-Implicit Connection}
\label{sec:dense}

The view that the dense connections among the different layers can improve the performance is verified in previous work \cite{huang2017densely}. For the framework in Fig.~\ref{fig:diaframwork}, the DIA unit also plays a role in bridging the current layer and the preceding layers such that the DIA unit can adaptively learn the non-linearity relationship between features in two different dimensions. Instead of explicitly dense connections, the DIA unit can implicitly link layers at different depths via a shared module which leads to dense connections.

\begin{figure}[t]
	 		\centering
	 		\includegraphics[width=1.0\linewidth]{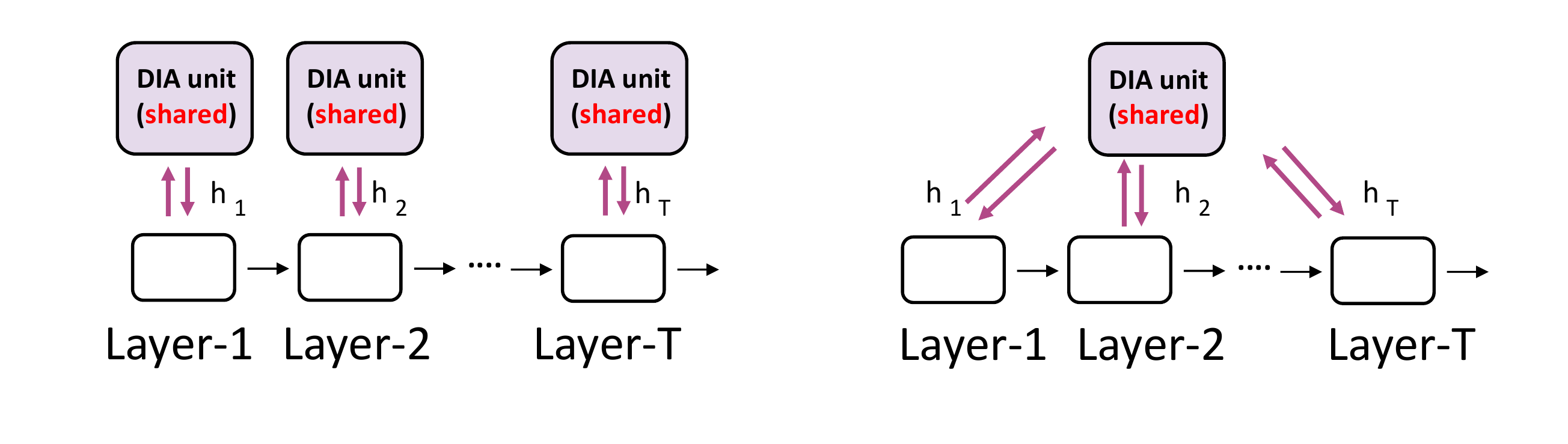}
	 		\caption{\textbf{Left.} Explicit structure of the network with DIA unit. \textbf{Right.} Implicit connection caused by DIA unit.} 
	 \vspace{-0.2cm}
	 		\label{fig:imp}
	 	\end{figure}

	  Specifically, consider a stage consisting of many layers in Fig.~\ref{fig:imp} (Left). It is an explicit structure with a DIA unit, and one layer seems not to connect to the other layers except the network backbone. In fact, the different layers use the parameter-sharing attention module, and the layer-wise information jointly influences the update of learnable parameters in the module, which causes implicit connections between layers with the help of the shared DIA unit as in Fig.~\ref{fig:imp} (Right). Since there is communication between every pair of layers, the connections over all layers are dense.  
	 	
	\begin{table}[ht]
			\centering
    \caption{The test accuracy (\%) when removing of DIA-LSTM in different stages of the ResNet backbone.}
			\begin{tabular}{ccccc}
				\toprule
				Stage removed & $\#$P(M) & $\#$P(M)$\downarrow$ & top1-acc. & top1-acc.$\downarrow$ \\
				\midrule
				stage1 & 1.94  & 0.01  & 76.85 & 0.24 \\
				stage2 & 1.90   & 0.05  & 76.87 & 0.22 \\
				stage3 & 1.78  & 0.17  & 75.78  & 1.31 \\
				\bottomrule
			\end{tabular}%

    \label{tab:remove stage}
\end{table}
\begin{table}[ht]
		\centering
		\caption{Testing accuracy (\%). We train models of different depths without BN on CIFAR-100. ``nan'' indicates the numerical explosion. }
  \resizebox*{0.99\linewidth}{!}{
		\begin{tabular}{lcccccc}
			\toprule
			 & \multicolumn{2}{c}{Origin} & \multicolumn{2}{c}{SENet} & \multicolumn{2}{c}{DIA-LSTM $(r=16)$} \\
			\cmidrule{2-7}        Backbone  & $\#$P(M) & top1-acc. & $\#$P(M) & top1-acc. & $\#$P(M) & top1-acc. \\
			\midrule
			ResNet83 & 0.88  & nan   & 0.98  & nan   & 0.94  & \textbf{72.34 } \\
			ResNet164 & 1.70  & nan   & 1.91  & nan   & 1.76  & \textbf{73.77 } \\
			ResNet245 & 2.53  & nan   & 2.83  & nan   & 2.58  & \textbf{73.86 } \\
			ResNet326 & 3.35  & nan   & 3.75  & nan   & 3.41  & nan \\
			\bottomrule
		\end{tabular}%
}
		\label{tab:withoutbn}%
	\end{table}%

	Next, we can try to further understand the dense connection from the numerical perspective. As shown in Fig.~\ref{fig:diaunit} and~\ref{fig:imp}, the DIA-LSTM bridges the connections between layers by propagating the information forward through $h_t$ and $c_t$, leading to feature integration. 
	Moreover, $h_t$ at different layers are also integrating with $h_{t'},1\leq t'<t$ in DIA-LSTM. Notably, $h_t$ is applied directly to the features in the network at each layer $t$. Therefore the relationship between $h_t$ at different layers somehow reflects the layer-wise connection degree. We explore the nonlinear relationship between the hidden state $h_t$ of DIA-LSTM and the preceding hidden state $h_{t-1},h_{t-2},...,h_1$, and visualize how the information coming from $h_{t-1},h_{t-2},...,h_1$ contributes to $h_t$. To reveal this relationship, we consider using the random forest to visualize variable importance. The random forest can return the contributions of input variables to the output separately in the form of importance measure, e.g., Gini importance~\cite{Gregorutti2017Correlation,huang2020dianet}. 

	Specifically, take $h_n,1\leq n < t$ as input variables and $h_t$ as output variable, we can get the Gini importance of each variable $h_n,1\leq n < t$. ResNet164 contains three stages, and each stage consists of 18 layers. We conduct three Gini importance computations for each stage separately. As shown in Fig.~\ref{features integration for each stage}, each row presents the importance of source layers $h_n,1\leq n < t$ contributing to the target layer $h_t$. In each sub-graph of Fig.~\ref{features integration for each stage}, the diversity of variable importance distribution indicates that the current layer uses the information of the preceding layers. The interaction between shallow and deep layers in the same stage reveals the effect of implicitly dense connection. In particular, for the results of DIA-LSTM, taking $h_{17}$ in stage 1~(the last row) as an example, $h_{16}$ or $h_{15}$ does not intuitively provide the most information for $h_{17}$, but $h_5$ does. The DIA unit can adaptively integrate information between multiple layers. 
	Unlike the DIA-LSTM, the information in each layer of the SENet tends to be strongly correlated with the information in the previous one or two layers, i.e., the Gini importance in the diagonal is significantly larger than that in the other positions.
		Moreover, in Fig.~\ref{features integration for each stage}~(stage 3) for DIA-LSTM, the information interaction with previous layers in stage 3 is more intense and frequent than that of the first two stages. Correspondingly, as shown in Table~\ref{tab:remove stage}, in the experiments, when we remove the DIA-LSTM unit in stage 3, the classification accuracy decreases from 77.09 to 76.85. However, when it is removed in stage 1 or 2, the performance degradation is very similar, falling to 76.85 and 76.87, respectively. Also note that for DIANet, the number of parameter increments in stage 2 is larger than that of stage 1. It implies that the significant performance degradation after the removal of stage 3 may be not only due to the reduction of the number of parameters but due to the lack of dense feature integration.
	
		 	\begin{figure*}[t]
	 		\centering
	 		\includegraphics[width=1.0\textwidth]{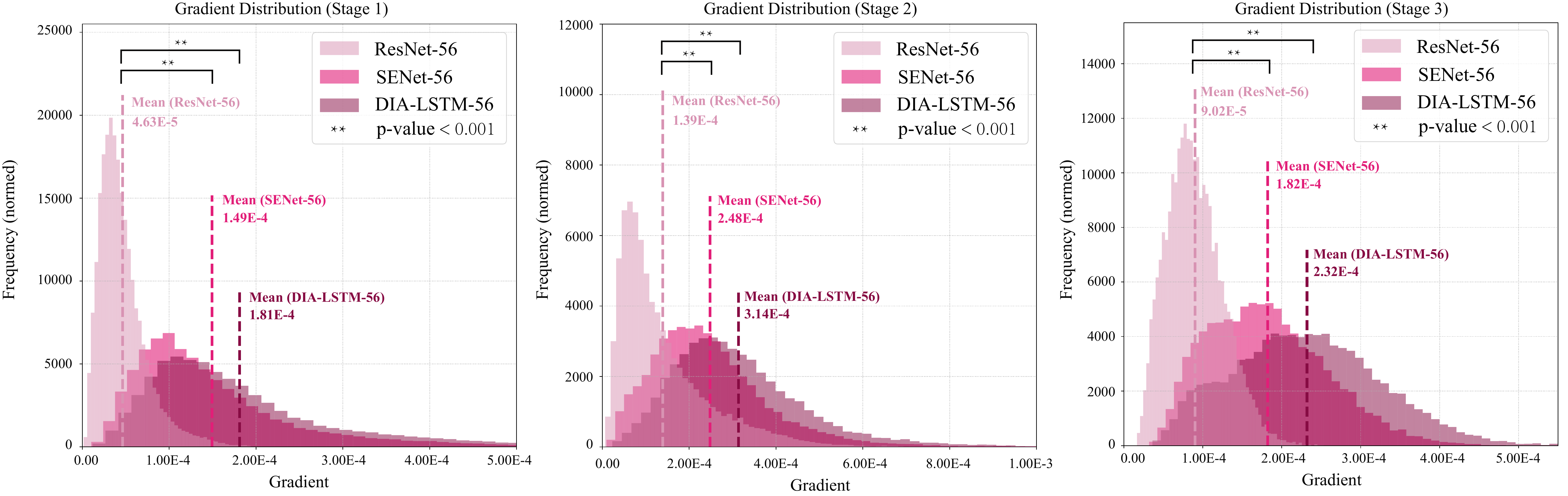}
	 		\caption{The distribution of gradient in each stage of ResNet56 without all the skip connections.} 
	 \vspace{-0.2cm}
	 		\label{fig:noskip_grad}
	 	\end{figure*}

	 	\begin{figure*}[t]
	 		\centering
	 		\includegraphics[width=0.75\textwidth]{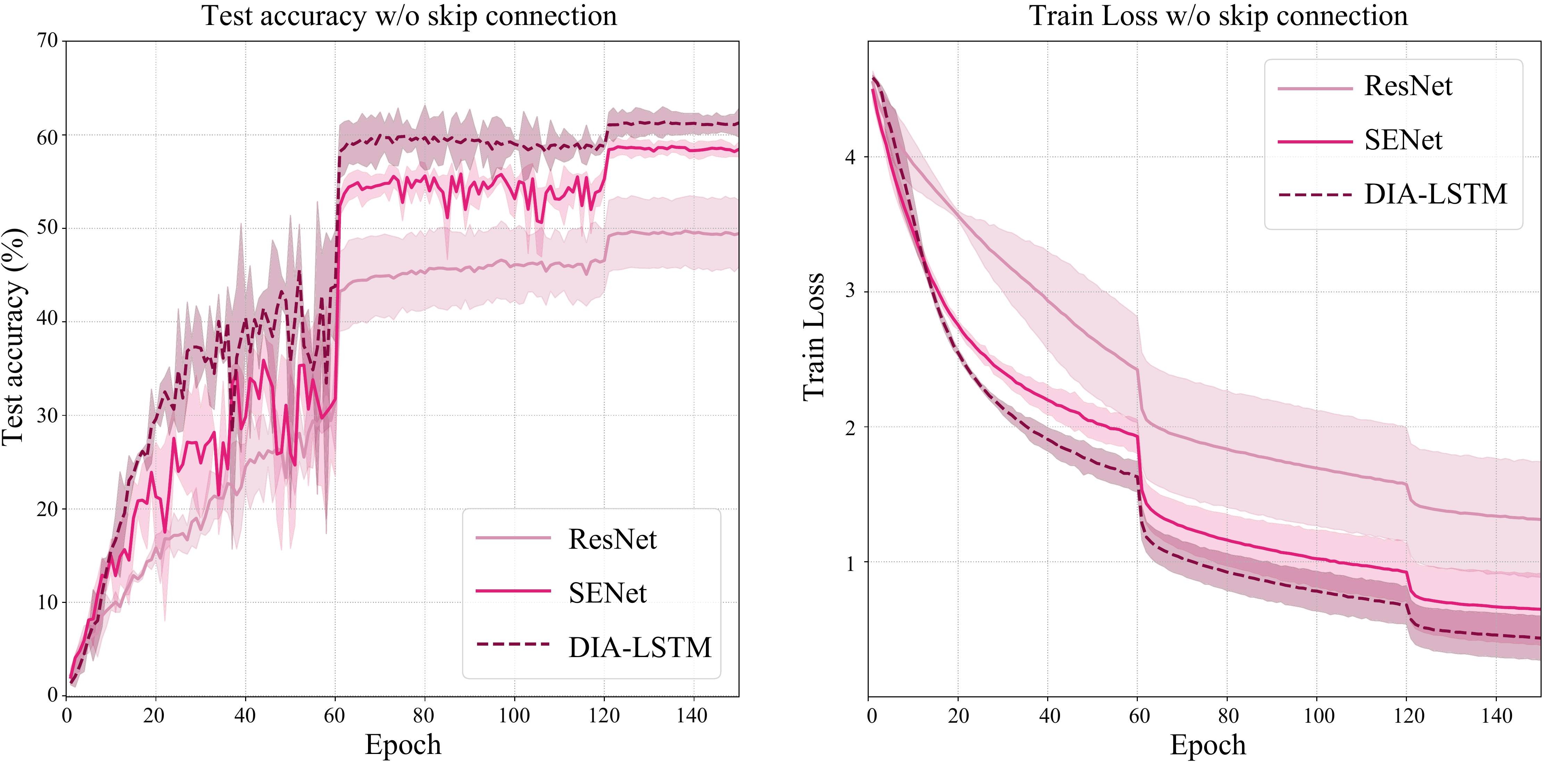}
	 		\caption{The test accuracy and train loss of ResNet164 without skip connection.} 
	 \vspace{-0.2cm}
	 		\label{fig:noskip_acc}
	 	\end{figure*}

		\begin{table}
			\centering
		\caption{Test accuracy (\%) of the models without data augment with ResNet164.}
			\begin{tabular}{ccc}
				\toprule
				Models    & CIFAR-10  & CIFAR-100 \\
				\midrule
				ResNet164 & 87.16 &61.42 \\
				SE & 87.21     & 63.92 \\
				DIA-LSTM & \textbf{89.42}     & \textbf{67.86} \\
				\bottomrule
			\end{tabular}%

		\label{tab:remove stage and without aug}%
    \end{table}{}	

\subsubsection{Stabilizing Training with DIA-LSTM}
\label{sec:stab}

In Section \ref{sec:dense}, we discussed the presence of dense-and-implicit connections. In this section, we present a series of experiments to showcase the regularization effect of these connections in stabilizing the training process.

\noindent\textbf{Removal of Batch Normalization.}
Small changes in shallower hidden layers may be amplified as the information propagates within the deep architecture and sometimes result in a numerical explosion. Batch Normalization (BN)~\cite{Ioffe:2015:BNA:3045118.3045167} is widely used in deep networks since it stabilizes the training by normalization the input of each layer. DIA-LSTM unit recalibrates the feature maps by channel-wise multiplication, which plays a role of scaling similar to BN. Table~\ref{tab:withoutbn} shows the performance of the models of different depths trained on CIFAR100, where BNs are removed in these networks. The experiments are conducted on a single GPU with a batch size of 128 and an initial learning rate of 0.1. Both the original ResNet and SENet face the problem of numerical explosion without BN, while the DIA-LSTM can be trained with a depth up to 245. In Table~\ref{tab:withoutbn}, at the same depth, SENet has a larger number of parameters than DIA-LSTM but still comes to numerical explosion without BN, which means that the number of parameters is not the case for stabilization of training, but sharing mechanism we proposed may be the case. Besides, compared with Table~\ref{tab:DIANet with different depth1}, the testing accuracy of DIA-LSTM without BN still can keep up to 70\%. It suggests that the scaling learned by DIA-LSTM integrates the information from preceding layers and enables the network to choose a better scaling for features of the current layer.

\noindent\textbf{Without Data Augmentation.}
Explicit dense connections may help bring more efficient usage of parameters, which makes the neural network less prone to overfit~\cite{huang2017densely}. Although the dense connections in DIA-LSTM are implicit, our method still shows the ability to reduce overfitting. To verify it, We train the models without data augment to reduce the influence of regularization from data augment. As shown in Table~\ref{tab:remove stage and without aug}, ResNet164 with DIA-LSMT achieves lower testing errors than the original ResNet164 and SENet. To some extent, the implicit and dense structure of our method may have a regularization effect.

\noindent\textbf{Removal of Skip Connection. }
The skip connection has become an essential structure for training DNNs~\cite{he2016identity}. Without skip connections, it becomes challenging to train the DNN due to issues such as gradient vanishing\cite{bengio1994learning,glorot2010understanding,srivastava2015training}. To investigate this further, we conducted an experiment where we removed all skip connections in ResNet56 and calculated the absolute value of the gradient at the output tensor of each stage.

In Figure~\ref{fig:noskip_grad}, we present the gradient distribution after removing all skip connections. We observe that when using the DIA-LSTM with ResNet56, the mean and variance of the gradient distribution noticeably increase. This increase allows for larger absolute values and greater diversity of gradients, thereby alleviating gradient degradation to some extent.

Furthermore, we can explore the ability of DIA-LSTM by examining the test accuracy and train loss when the model is trained without skip connections. We adopt the training configuration described in \cite{huang2020dianet} to train the ResNet with various self-attention modules and present the test accuracy and train loss in Figure~\ref{fig:noskip_acc}.

In terms of test accuracy, we observe that skip connections play a crucial role in the training process. When skip connections are not utilized, the model's performance suffers a significant decline, with classification accuracy dropping to approximately 50\% to 65\% for all models. Furthermore, the test accuracy curves exhibit oscillations, indicating that training the model becomes unstable without skip connections. In this scenario, DIA-LSTM proves to be effective in mitigating the performance loss caused by the absence of skip connections when compared to the original ResNet and SENet architectures. Likewise, when considering train loss, DIA-LSTM consistently demonstrates the fastest rate of optimization. These two experiments suggest that DIA-LSTM can serve as an effective regularizer for training neural networks.

\subsection{Discussion}
\label{sec:limit}
In this section, we will discuss two potential issues in terms of computational cost and the sharing strategy.
\subsubsection{Computational cost}
As demonstrated in our extensive experiments in Section \ref{sec:exp}, the incorporation of LSTM brings substantial benefits to our proposed framework. However, it is important to note that employing LSTM may result in increased computational costs and longer training times. For instance, when considering ResNet164, the inclusion of a standard DIA-LSTM module leads to an approximately 12\% slower inference speed compared to the absence of any self-attention module.

Consequently, in scenarios where computational speed is a crucial factor, such as applications with sensitivity to real-time processing, it is advisable to utilize a more efficient LSTM design, as depicted in Figure \ref{fig:lstm}. This alternative design, referred to as DIA-LSTM (Light), manages to mitigate the aforementioned issue while incurring a minor performance loss. Notably, DIA-LSTM (Light) reduces the computational cost bottleneck from 12\% to a mere 1\%.

	 	\begin{figure}[t]
	 		\centering
	 		\includegraphics[width=1.0\linewidth]{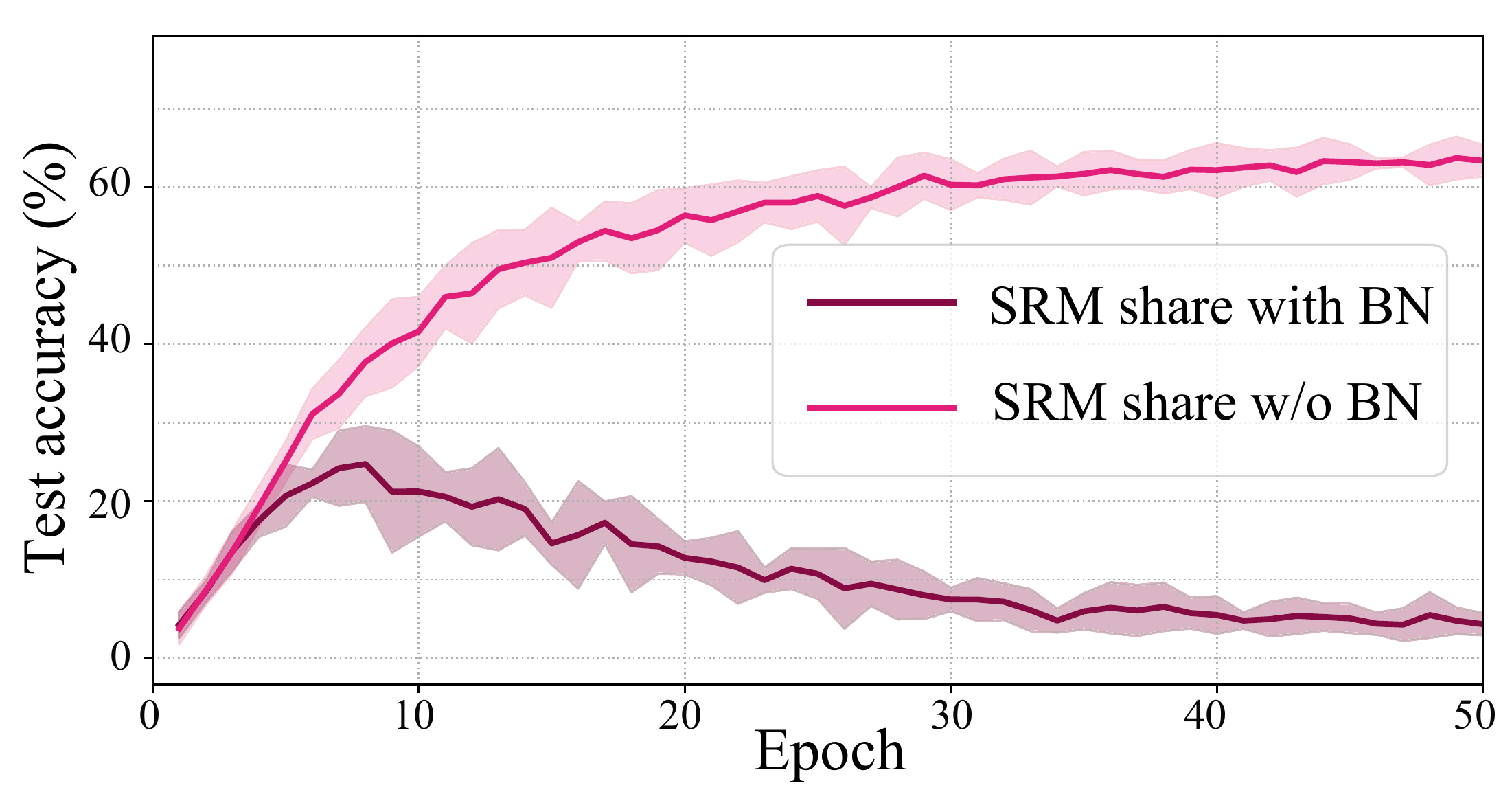}
	 		\caption{The performance of SRM with two sharing strategies. } 
	 \vspace{-0.2cm}
	 		\label{fig:withbn}
	 	\end{figure}

\subsubsection{Incorporating batch normalization with DIA}

While our proposed framework, illustrated in Figure \ref{fig:diaframwork}, offers enhancements for various self-attention modules, it is important to note that not all of these modules can be directly shared across the backbone layer. Specifically, as indicated in Table \ref{tab:diaformodule}, when applying SRM to the DIA unit, batch normalization (BN) should not be shared across layers.

Indeed, batch normalization (BN) is a commonly employed component in deep learning architectures, including those incorporating self-attention mechanisms \cite{lee2019srm,liang2020instance}. However, previous studies \cite{wang2021recurrent,stelzer2021deep,Jaiswal2021TDAMTA} have revealed that the statistics computed by BN are closely tied to the characteristics of the specific layer where it is located. Consequently, for certain tasks, it is not recommended to utilize these statistics across different layers. To validate our statement, we conducted image classification experiments on the CIFAR100 dataset using the ResNet164 backbone. Specifically, we examined the training performance when incorporating SRM with the DIA unit and compared the results obtained by sharing or unsharing the BN parameters. Figure~\ref{fig:withbn} displays the classification results for the initial 50 epochs using two distinct sharing strategies for the SRM module. We observe that when BN is shared, the model's classification performance fluctuates during the early training stages, eventually plummeting to an accuracy level near 0. Conversely, when BN is not involved in sharing, the model's classification accuracy steadily improves, aligning with our analysis.

\section{Conclusion}
\label{conculsion}
In this paper, \hzz{inspired by a counterintuitive but inherent observation, i.e., SAMs tend to produce strongly correlated attention maps across different layers, we propose a generic shared attention mechanism called the DIA to improve the parameters' utilization of SAMs.} We show the effectiveness of the  proposed mechanism for several popular vision tasks by conducting experiments on benchmark datasets and popular networks. Moreover, we further empirically demonstrate that the dense and implicit connections formed by our method have a strong regularization ability to stabilize the training of DNNs.

	\bibliographystyle{IEEEtran}
	\typeout{}
	\bibliography{ref2.bib}

\vfill

\end{document}